\documentclass[times,twocolumn,final]{elsarticle}

\usepackage{adjustbox}
\usepackage{algorithmic}
\usepackage{amsfonts}
\usepackage{amsmath}
\usepackage{amssymb}
\usepackage{array}
\usepackage{arydshln}
\usepackage{booktabs}
\usepackage{cite}
\usepackage{color}
\usepackage{comment}
\usepackage{etoolbox}
\usepackage{framed}
\usepackage{graphicx}
\usepackage{hyperref}
\usepackage{ifthen}
\usepackage{latexsym}
\usepackage{makecell}
\usepackage{medima-arxiv}
\usepackage{multicol}
\usepackage{multirow}
\usepackage{pifont}
\usepackage{tabularx}
\usepackage{tabulary}
\usepackage{textcomp}
\usepackage{url}

\definecolor{newcolor}{rgb}{.8,.349,.1}

\journal{Medical Image Analysis}

\def\rot{\rotatebox}
\newcommand{\used}{\ding{51}}

\newcommand{\citeb}[1]{}
\newcommand{\videolink}{\url{https://vimeo.com/951853260}}
\newcommand\blfootnote[1]{%
  \begingroup
  \renewcommand\thefootnote{}\footnote{#1}%
  \addtocounter{footnote}{-1}%
  \endgroup
}

\begin{document}

\verso{Chinedu Nwoye \textit{et~al.}}

\begin{frontmatter}

\title{SurgiTrack: Fine-grained multi-class multi-tool tracking in surgical videos}%

\author[1,2]{Chinedu Innocent Nwoye\corref{cor1}}
\cortext[cor1]{Corresponding author}
\ead{nwoye@unistra.fr}
\author[1,2]{Nicolas Padoy}

\address[1]{University of Strasbourg, CAMMA, ICube, CNRS, INSERM, France}
\address[2]{IHU Strasbourg, Strasbourg, France}

\begin{abstract}
{\small
Accurate tool tracking is essential for the success of computer-assisted intervention.
Previous efforts often modeled tool trajectories rigidly, overlooking the dynamic nature of surgical procedures, especially tracking scenarios like out-of-body and out-of-camera views. 
Addressing this limitation, the new CholecTrack20 dataset provides detailed labels that account for multiple tool trajectories in three \textit{perspectives}: (1) intraoperative, (2) intracorporeal, and (3) visibility, representing the different types of temporal duration of tool tracks. 
These fine-grained labels enhance tracking flexibility but also increase the task complexity. Re-identifying tools after occlusion or re-insertion into the body remains challenging due to high visual similarity, especially among tools of the same category.
This work recognizes the critical role of the tool operators in distinguishing tool track instances, especially those belonging to the same tool category. The operators' information are however not explicitly captured in surgical videos. We therefore propose SurgiTrack, a novel deep learning method that leverages YOLOv7 for precise tool detection and employs an attention mechanism to model the originating direction of the tools, as a proxy to their operators, for tool re-identification. To handle diverse tool trajectory perspectives, SurgiTrack employs a harmonizing bipartite matching graph, minimizing conflicts and ensuring accurate tool identity association.
Experimental results on CholecTrack20 demonstrate SurgiTrack's effectiveness, outperforming baselines and state-of-the-art methods with real-time inference capability. This work sets a new standard in surgical tool tracking, providing dynamic trajectories for more adaptable and precise assistance in minimally invasive surgeries.
}
\end{abstract}


\end{frontmatter}



\section{Introduction}
\label{sec:introduction}

\begin{figure}[!t]
    \centering
    \includegraphics[width=0.99\columnwidth]{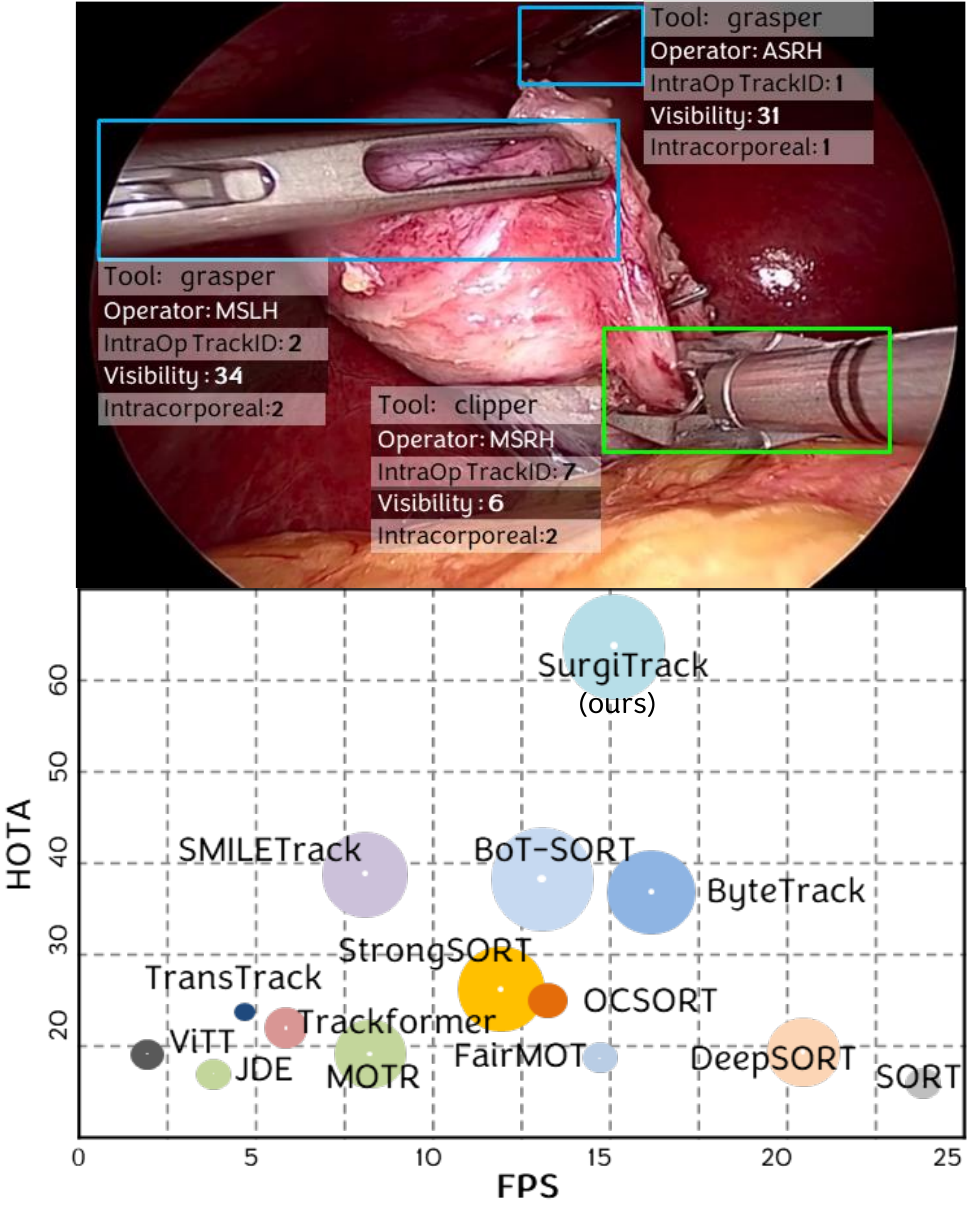}
    \caption{{Surgical tool tracking demonstrating (\textit{top}) qualitative fine-grained tracking result across multiple tools, classes, and perspectives and (\textit{bottom}) superior quantitative results compared to the state-of-the-art.}}
    \label{fig:featured-img}
\end{figure}

\blfootnote{Article to appear in Medical Image Analysis Journal, 2025}
Surgical tool tracking plays a pivotal role in computer-assisted surgical systems, offering valuable insights for a range of applications, including skill assessment \citep{pedrett2023technical}, visual servoing \citep{xu2023graph}, navigation \citep{xu2022information}, laparoscope positioning \citep{dutkiewicz2005experimental}, safety and risk zone estimation \citep{richa2011visual}, and augmented reality \citep{martin2023sttar}.
While tool detection identifies target tools in an image or frame, tool tracking goes a step further to also include the estimation and prediction of the tools' locations as they appear in subsequent video frames.
Historically, tool tracking relied on traditional machine learning features, encompassing color, texture, SIFT, and geometry \citep{pezzementi2009articulated,sznitman2012data,alsheakhali2015surgical,dockter2014fast,du2016combined}. Recent advances in deep learning \citep{bouget2017vision,lee2019weakly,nwoye2019weakly,zhao2019surgical,zhao2019real,robu2021towards,nwoye2021deep,fathollahi2022video,wang2022visual,rueckert2023methods} ushered in a new era by enabling the extraction of more robust features for tool re-identification (re-ID).
Despite the remarkable progress, challenges remains.
Existing efforts have primarily focused on single tool tracking \citep{zhao2019surgical}, single-class multi-tool tracking \citep{fathollahi2022video}, or multi-class single-tool tracking \citep{nwoye2019weakly}. 
Whereas in real-world surgical scenarios, 
multiple tools of varying classes are often utilized simultaneously, necessitating multi-class multi-tool tracking, a domain that remains largely unexplored mainly due to lack of the requisite dataset.

Recently, a new dataset known as CholecTrack20 \citep{Nwoye2023CholecTrack20} was introduced providing the multi-class multi-tool tracking requirements. This dataset also formalized three different \textit{perspectives} of trajectories capturing (1) the life-long intraoperative use of tools, (2) intracorporeal cycle of the tools within the body, and (3) the visibility lifespan of the tools within the camera field of view (FoV) as shown in Fig. \ref{fig:featured-img}. 
Simultaneously tracking tools across these three perspectives is termed \textit{multi-perspective tracking}.
The CholecTrack20 dataset provides rich multi-perspective tracking annotations adaptable to diverse surgical needs, however, to date, no deep learning model has been explored on this dataset for automatic surgical tool tracking.

To develop a method for multi-perspective multi-class multi-tool tracking in surgical videos, we first benchmark 10 state-of-the-art (SOTA) detection methods on the CholecTrack20 dataset and conduct an extensive ablation study on suitable re-ID methods for tracking in the surgical domain.
The re-ID module plays a pivotal role of managing the tools identities across time in surgical videos. But, challenges arise from the intricate motion patterns of the tools, frequent occlusions, and the limited field of view within the surgical scene. A particularly daunting task is the re-identification of tools after they have been occluded, moved out of the camera's view, or re-inserted into the surgical field. This complexity is amplified when multiple instances of the same tool class share identical appearance features.
Contrary to existing approaches, our preliminary experiments revealed that relying solely on tool appearance cues for track discrimination is sub-optimal especially when distinguishing between instances of the same class. To address this issue, we turn to domain knowledge, specifically the tool's usage pattern and the tool operator's information. The latter criterion, tool operator, refers to the surgeon's hand manipulating the tool and is found to be more accurate than appearance features in distinguishing between instances of the same tool class. However, the tool operators are not directly observable in endoscopic images, making their automatic prediction a challenging endeavor.

Inspired by our findings, we propose {\it SurgiTrack}, a novel deep learning approach for surgical tool tracking.
SurgiTrack casts the tool operators as the approximation of the tools' originating directions.
It then employs an Attention Mechanism to encode tool motion direction, effectively emulating the unseen surgeon operator's hands or trocar placements for tool re-identification.
Our model design allows for an alternative self-supervision of the direction estimator with a comparable performance to its supervised counterpart. This technique ensures that our method can be explored on datasets where the tool operator labels are unavailable.
Finally, to account for the multi-perspective nature of tool trajectories, our network associate tracks using a \textit{harmonizing} bipartite matching graph algorithm, which, aside from the usual linear assignment, resolves identity conflicts across the track perspectives and improves the accuracy of the track ID re-assignment in general. 

To summarize, our contributions with this work include the formalization of multi-perspective tool tracking modeling and benchmarking of state of the art methods on the CholecTrack20 dataset.
It also includes the development of the SurgiTrack model which relies on self-supervised attention-based motion direction estimation and harmonizing bipartite graph matching for tool tracking.
Finally, we provided an extensive evaluation of tool tracking on the different trajectory perspectives, at varying video frame rates, and under various visual challenges such as bleeding, smokes, occlusion, etc. 
These contributions collectively advance the state of the art in surgical tool tracking, opening doors to enhanced computer-assisted surgical systems and AI interventions in the field.

\begin{figure*}[!th]
    \centering
    \includegraphics[width=.99\linewidth]{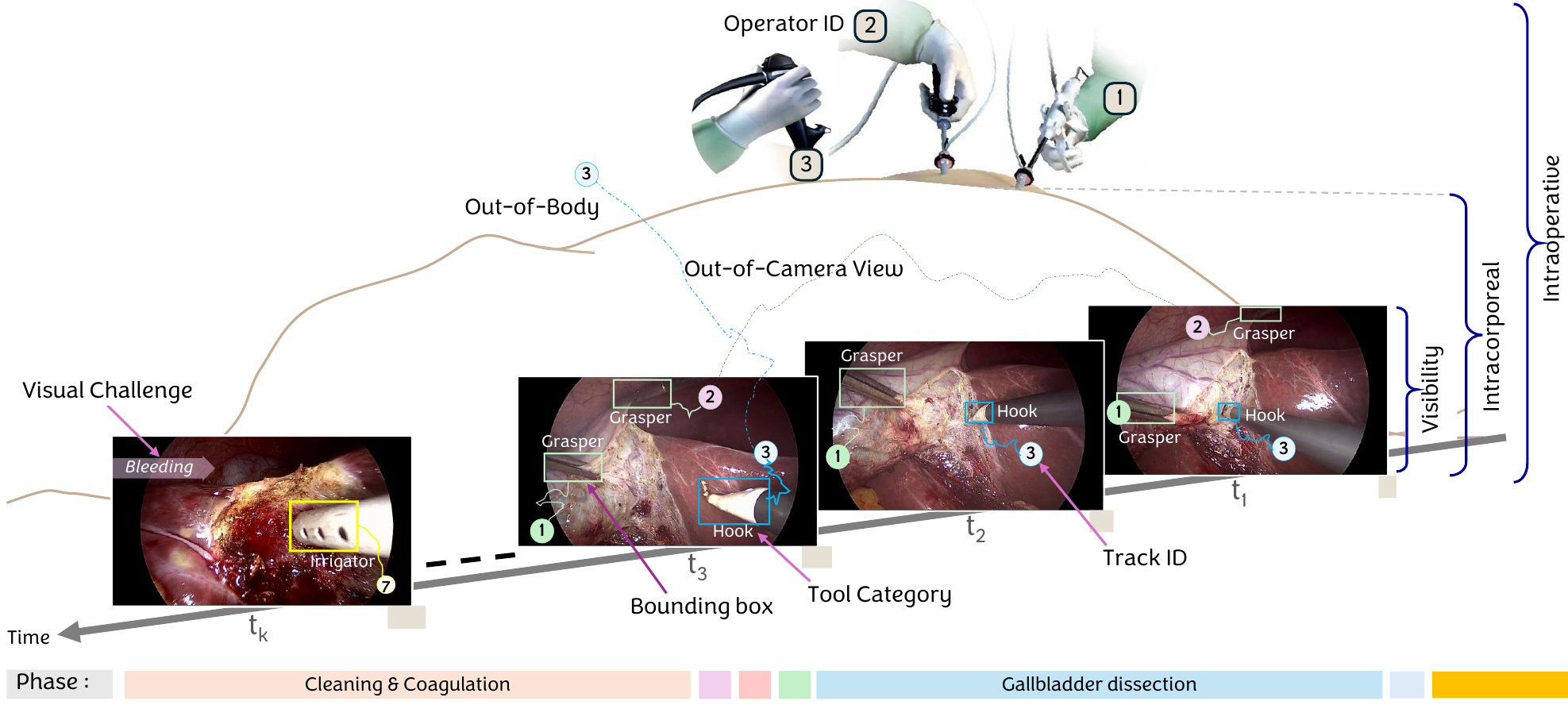}
    \caption{{ Overview of CholecTrack20 dataset showing localization, tracking, and associated labels \citep{Nwoye2023CholecTrack20}.}}
    \label{fig:data-overview}
\end{figure*}

\begin{table*}[!thpb]
    \centering
    \caption{{Summary of the CholecTrack20 dataset statistics \citep{Nwoye2023CholecTrack20}.}}
    \label{tab:dataset-stats}
    \setlength{\tabcolsep}{4pt}
    \resizebox{0.95\linewidth}{!}{%
    \begin{tabular}{@{}lcrlcccrlc@{}}
    \toprule 
        \multicolumn{2}{c}{Details} &\phantom{abc}& 
        \multicolumn{4}{c}{Average track length / number of tracks per perspective} &\phantom{abc}& \multicolumn{2}{c}{Surgical visual challenges}\\
        \cmidrule{1-2}\cmidrule{4-7}\cmidrule{9-10}
        Params & Counts && Tools & Intraoperative & Intracorporeal & Visibility && Events & Counts  \\
        \midrule\midrule
        Videos & 20 && Grasper & 582.0 / 59 & 209.4 / 164 & 25.7 / 1334 && Occlusion & 22958 \\
        Total duration & 14h 3m 1s && Bipolar & 118.0 / 20 & 30.6 / 78 & 24.6 / 96 && Bleeding & 18698 \\
        Tool categories & 7  && Hook & 1063.7 / 20 & 181.8 / 118 & 87.5 / 244 && Smoke & 2388 \\
        Phase categories & 7  && Scissors & 44.6 / 20 & 27.0 / 34 & 23.4 / 46 && Crowded & 6737 \\
        Operator categories & 4  && Clipper & 71.7 / 20 & 36.8 / 40 & 31.9 / 46 && Blurring & 299 \\
        Labeled frames (1 FPS) & 35K  && Irrigator & 90.8 / 20 & 25.9 / 56 & 13.4 / 108 && Light reflection & 74 \\
        Total frames (25 FPS) & 1.3M && Spec.Bag & 175.2 / 20 & 166.8 / 20 & 44.9 / 78 && Fouled lens & 2196 \\
        Bounding boxes & 65.2K  && \textit{Total} & 2146.0 / 179 & 678.3 / 510 & 251.4 / 1952 && Poor camera coverage & 357 \\
        \bottomrule        
    \end{tabular}
    }
\end{table*}

\section{Related Work}
\label{sec:literature}
Despite the many potentials of tool tracking, challenges persist, prompting our research into more robust and efficient tracking techniques.
In the domain of surgical tool tracking, an array of tracking modalities \citep{fried1997image,chmarra2007systems,song2023arc,speidel2008recognition,Behrens2011Inertial,reiter2012feature,sznitman2012data}, ranging from electromagnetic to optical and image-based tracking, has been explored to enhance precision and efficiency. While robotic kinematics gains popularity for its accuracy in instrument localization, image-based tracking stands out for its non-invasiveness, applicability, generalizability, and seamless alignment with the surgeon's visuals. Despite its merits, image-based tracking faces challenges such as occlusion, deformation, and lighting issues. Our work aligns with the trajectory of existing deep learning research \citep{
bouget2017vision,lee2019weakly,nwoye2019weakly,zhao2019surgical,zhao2019real,robu2021towards,nwoye2021deep,fathollahi2022video,wang2022visual,rueckert2023methods}, contributing to the ongoing refinement of image-based tracking, positioning it as a pivotal frontier in revolutionizing surgical assistance.

The dynamic landscape of tool tracking extends beyond single-object tracking (SOT) \citep{danelljan2017eco,zhao2019surgical} to encompass multi-object tracking (MOT) \citep{wojke2017deepsort,wang2020towards,zhang2021fairmot,robu2021towards} and multi-class tracking (MCT) \citep{nwoye2019weakly,nwoye2021deep}. For diverse surgical tools, our research lies within the scope of multi-class multi-object tracking (MCMOT) \citep{lee2016multi,jo2017real,zhu2021detection,du2021giaotracker}, reflecting its potential for real-world surgical applications \citep{pedrett2023technical,xu2023graph,xu2022information,dutkiewicz2005experimental,richa2011visual,martin2023sttar}. In the pursuit of robust tracking methodologies, the choice between tracking paradigms, such as tracking by detection \citep{zhang2021fairmot,wang2020towards,aharon2022bot,zhang2022bytetrack,wang2023smiletrack}, regression \citep{bergmann2019tracking}, attention \citep{sun2020transtrack,zeng2022motr,meinhardt2022trackformer,chu2023transmot}, segmentation \citep{lee2019weakly} or fusion \citep{reiter2012feature}, guides the selection based on context and task suitability. Our research adopts a tracking-by-detection paradigm for its simplicity in achieving surgical tool tracking precision.

Amidst the evolution of tracking methods from rule-based and marker-based approaches \citep{ma2021surgical,huang2020tracking,cartucho2022enhanced} to markerless techniques \citep{reiter2012feature,ye2016real} and machine learning methods \citep{rieke2015surgical,zhao2019surgical,alsheakhali2015surgical,pezzementi2009articulated,speidel2008recognition,dockter2014fast}, our work is positioned at the forefront, leveraging deep learning advancements like Siamese networks and attention-based networks \citep{sun2020transtrack,meinhardt2022trackformer}. Notably, our focus on direction features, capturing the tool handling direction by surgeon operators, sets our approach apart, offering a more suitable discriminator for refined tool tracking accuracy in scenarios where appearance \citep{reiter2014appearance} or similarity \citep{wang2023smiletrack} features may prove less robust. 
Contrary to the joint detection and embedding (JDE) setting \citep{zhang2021fairmot,wang2020towards}, adopting an online separate detection and tracking (SDE) approach \citep{bewley2016simple,maggiolino2023deep,du2023strongsort,aharon2022bot,du2021giaotracker,wang2023smiletrack}, where detection precedes tracking, aligns with the essence of an independently trained tracker that assumes prior detection results.
Anchored in dataset like CholecTrack20 \citep{Nwoye2023CholecTrack20}, which offers both multi-perspective trajectories and MCMOT requirements in endoscopic videos, our research aligns with the ongoing efforts \citep{nwoye2021deep,robu2021towards,fathollahi2022video,rueckert2023methods} to address the complexities of surgical tool tracking, contributing to the broader landscape of advancements in this critical domain.

\begin{figure*}[!t]
    \centering
    \includegraphics[width=0.99\textwidth]{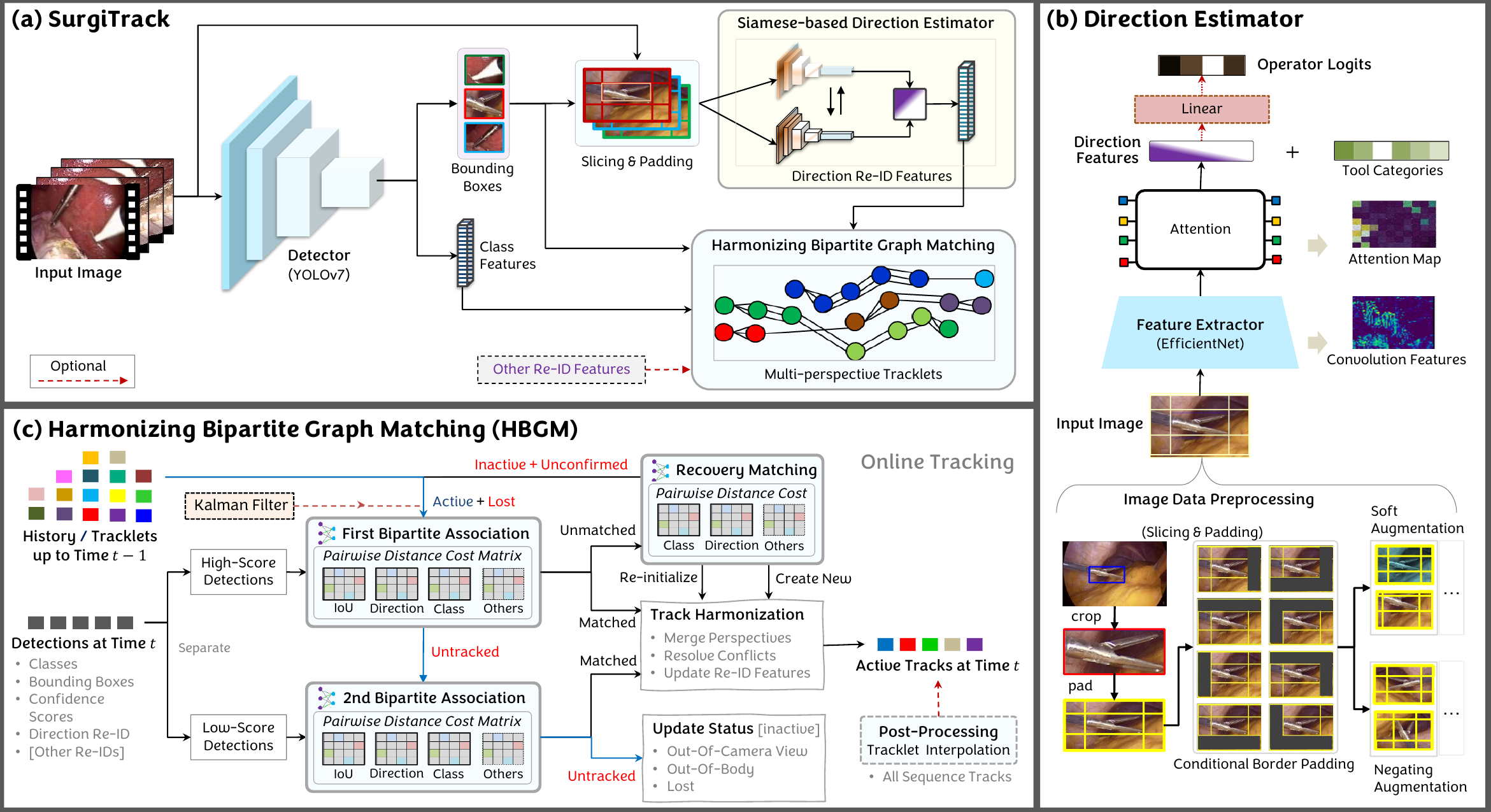}
    \caption{Overview of our proposed tool tracking model showing: (a) full architecture of SurgiTrack and its major component modules. One of the which is the YOLO-based detector. 
    The other is the Siamese-based surgical tool direction estimator - full architectural detail in (b) which also shows an optional head for surgeon operator classification. The last component of the SurgiTrack is the harmonizing bipartite graph matching (HBGM) algorithm for tool track identity association under multiple perspectives of tool trajectories: visibility, intracorporeal, and intraoperative - full pipeline in (c).
    }
    \label{fig:architecture}
\end{figure*}


\section{Dataset} 
\label{sec:data}

Our research is conducted on CholecTrack20  \citep{Nwoye2023CholecTrack20} dataset consisting of 20 videos of laparoscopic procedures that have been fully annotated with detailed labels for multi-class multi-tool tracking.
The dataset, illustrated in Fig. \ref{fig:data-overview}, provides track identities across 3 perspectives of track definition: (1) visibility trajectory of a tool within the camera scope, (2) intracorporeal trajectory of a tool while within a patient's body, and (3) life long intraoperative trajectory of a tool.
{Intraoperative tracking not only re-identifies tools out of camera view (OOCV) as done in intracorporeal tracking but also maintains their trajectory when out of body (OOB). In the CholecTrack20 dataset, OOB is detected/annotated either by visually observing the tool exit the trocar, inferring from another tool entering through the same trocar, or noting that the initial tool releases its grasp while out of camera focus.}
The dataset also provides detailed labels for each tool such as spatial bounding box coordinates, class identity, operator identity, phase identity, frame visual conditions such as occlusion, bleeding, and presence of smoke statuses, among others.
The annotated tool categories are grasper, bipolar, hook, scissors, clipper, irrigator and specimen bag. The annotated tool operators are main surgeon left hand (MSLH), main surgeon right hand (MSRH), assistant surgeon right hand (ASRH) and null operator (NULL). The annotations are provided at 1 frame per second (FPS) consisting of 35K frames and 65K instance tool labels. Raw videos, recorded at 25 FPS, are provided for inference. 
The dataset is available on Synapse \citep{dataRefNo}.
{A summary of the CholecTrack20 dataset statistics showing the size of key elements in the dataset, number of challenging scenarios, average track duration and number of tracks for each tool category per perspective is presented in Table \ref{tab:dataset-stats}.}
More information on the dataset can be found in \citet{Nwoye2023CholecTrack20}.


\section{Methods}

We present \textit{SurgiTrack}, a deep learning method for surgical tool tracking based on tool direction of motion features.
\textit{SurgiTrack} is designed as a multi-class multi-object tracking (MCMOT) model capable of tracking tools jointly across multiple trajectory perspectives, namely, visibility, intracorporeal, and intraoperative. The motivation to track beyond camera's field of view is to offer more flexible trajectories that ensure continuous and reliable identification of surgical tools, tailored to the complex dynamics of a surgical scene, preventing errors and maintaining safety even when tools temporarily move out of view.

The architecture of our proposed tracking model is conceptually divided into the main components of object tracking: spatial detection and data association, with the later further split into re-identification feature modeling and track identity matching, as illustrated in Fig. \ref{fig:architecture}(a).

\subsection{Multi-Perspective Tool Track Formalization}
Given a video dataset $ D=\{S_1, S_2,\ldots,S_N\} $, where each sequence $ S_i = \{f_1, f_2, \ldots, f_T\} $ consists of frames $ f_t $ containing tools represented by bounding box locations $ B_t = [B_1^t, B_2^t, \ldots, B_M^t] $ and associated classes $ C_t = [C_1^t, C_2^t, \ldots, C_M^t] $, the objective is to accurately track tool identities over time across multiple perspectives. For each perspective $ p \in \{1, 2, 3\} $, we define an association matrix $ A^{(p)}(t) $, where $ A^{(p)}_{i,j}(t) = 1 $ indicates that the $ i $-th detected tool in frame $ f_t $ is associated with the $ j $-th detected tool in frame $ f_{t+1} $, and $ A^{(p)}_{i,j}(t) = 0 $ otherwise. The final tracking solution involves harmonizing these perspective-specific matrices $ \{A^{(1)}(t), A^{(2)}(t), A^{(3)}(t)\} $ to ensure consistent and accurate tool tracking.

\subsection{Spatial Detection}
The spatial detection module is responsible for detecting the tools in each frame of a laparoscopic video. 
The detection output $D_t = (B_i^t, C_i^t)$ of frame $f_t$ is a pair of bounding box coordinates and class identities, with $B_i^t=[X_i^t, Y_i^t, W_i^t, H_i^t]$ containing the center coordinates $(X_i^t, Y_i^t)$ and the spatial sizes $(W_i^t, H_i^t)$ of the boxes and $C_i^t \in \{0, 1, \ldots, 6\}$ represents the class identity of the tools.

We use YOLOv7 \citep{wang2023yolov7}, a single shot object detector renowned for balancing accuracy and real-time inference, as the backbone for our detection module.
Pretrained on the Crowdhuman (CH) \citep{shao2018crowdhuman} and MOT20 datasets \citep{dendorfer2020mot20}, then finetune on the CholecTrack20 data, YOLOv7 produces tool class identities $C=[0,1, \ldots, 6]$ and bounding boxes $B=[X,Y,W,H]$, followed by Non-Maximum Suppression (NMS) at a 0.3 threshold to filter out redundant detections. The predicted bounding boxes are then used to crop out the tool regions from the RGB image for the Re-ID phase.

\subsection{Re-Identification by Direction Estimation}
Re-identification of tool instances is essential for managing their identities across time. It involves learning representations that can uniquely identify tool instances across frames.
In a video frame where multiple tools can share identical appearance, given that every tool is uniquely tied to a tool operator which remains consistent for their distinct trajectory, automatic estimation of these operators would be helpful in overcoming the challenges of tracking surgical tools in videos.
The tools operators, however, are not observable from the captured images.
We argue that the operator of a tool can be discerned from the originating direction of the tools termed the \textit{fro-direction}. 
We therefore propose a 3-step process of (1) data preprocessing, (2) direction estimation, and (3) operator prediction, as shown in Figure \ref{fig:architecture}(b), to indirectly infer the hidden operators from a given image by learning tool direction.

\subsubsection{Data Preprocessing}
We observed that the tool shaft, while not part of the bounding box annotations, more consistently points toward the fro-direction than the tip of the tool, which otherwise aligns with the to-direction. 
However, there is a limitation – the tool shaft may not always be visible in the images, especially when the tool is positioned near the corners of the frame.
Our preprocessing technique of {\textit{image slicing and padding}}, shown in Figure \ref{fig:architecture}(b), addresses this issue. Here, we crop the images using the tool bounding boxes and pad the cropped images with a percentage of the surrounding neighborhood pixels. Groundtruth and predicted bounding boxes are used at training and testing times respectively.
Where a bounding box is at the border of the frame, zero filled padding is applied. This approach helps to reveal the presence or absence of tool shafts.

\subsubsection{Direction Estimation}
We model the estimation of the direction $\theta$ of a surgical tool, given its bounding box location $B$, using a Bayesian-inspired framework. Specifically, the relationship between $\theta$ and $B$ follows Bayes' rule as shown in Equation \ref{eqn:bayes1}:
\begin{equation}
    P(\theta|B) = \frac{P(B|\theta) \cdot P(\theta)}{P(B)}
    \label{eqn:bayes1}
\end{equation}
Here, $P(B|\theta)$ represents the likelihood of the tool's bounding box given its direction, $P(\theta)$ is the prior probability of the direction, and $P(B)$ is the evidence, representing the marginal probability of the bounding box across all possible directions.

We approximate this framework using the Attention Mechanism ${\bf A}$ in Equation \ref{eqn:attn}, which captures dependencies between bounding box features and direction:
\begin{equation}
    {\bf A(Q, K, V)} = \text{\it SoftMax} \left( \frac{{\bf KQ}^T}{\sqrt{d_{\bf K}}} \right) {\bf V}
    \label{eqn:attn}
\end{equation}
The Attention Mechanism is a common approach in neural networks for learning dependencies between variables \citep{bahdanau2014neural}. In this formulation, the likelihood $P(B|\theta)$ is approximated by the attention alignment scores $\text{\it SoftMax}~({\bf KQ}^T / \sqrt{d_{\bf K}})$, which measure the relevance of each direction based on the bounding box features, i.e., how much attention (weight) each trocar direction should get based on the current tool location.
The prior probability $P(\theta)$ is represented by the value vector ${\bf V}$, which learns a distribution over possible directions independent of the bounding box features.
The evidence $P(B)$ serves as a normalization term, implicitly handled through the $\text{\it SoftMax}$ operation.
The query ${\bf Q}$, key ${\bf K}$, and value ${\bf V}$ are learned embeddings that encode direction vectors, bounding box features, and direction priors, respectively, from the input images using convolutional neural network (CNN) layers. Specifically, we use EfficientNet-b0 \citep{tan2019efficientnet} to encode these features.

The Attention Mechanism ${\bf A}$ effectively approximates $P(\theta|B)$ by learning the alignment between bounding box features and directional priors within a differentiable framework. Unlike classical Bayesian methods where priors and likelihoods are explicitly defined, this approach allows the network to learn these distributions directly from the data, providing flexibility for modeling complex spatial dependencies.
By grounding the formulation in Bayes' rule and leveraging the interpretability of attention mechanisms, we provide a principled approach to estimating tool direction while maintaining compatibility with modern deep learning architectures.

\subsubsection{Operator Estimation} 
The tool operator $O$ is estimated as the conditional probability $P(O \mid [{\bf A}, C])$, where ${\bf A}$ represents the attention-based direction features (derived from Equation \ref{eqn:attn}), and $C$ is the tool category.
This estimation is modeled using a fully connected layer, with the linear transformation defined as:
\begin{equation}
    y = \mathbf{w} \cdot [{\bf A}, C] + \mathbf{b}
    \label{eqn:lln}
\end{equation}  
where, $\mathbf{w}$ and $\mathbf{b}$ are the weight and bias parameters, and the input is the concatenation of ${\bf A}$ and $C$. 
To obtain the final probability distribution, a Softmax function is applied to $y$:

\begin{equation}
    P(O \mid [{\bf A}, C]) = \text{Softmax}(y)
\end{equation}  
This process generates the probability of each operator $O$ given the tool direction and category, enabling operator classification.

\subsubsection{Supervision} 
Our Direction Estimation Network can be trained through multiple paradigms, depending on the availability of labels. 
For \textit{full supervision}, the model is trained on and to predict the operator ID labels, employing a weighted cross-entropy loss for class balancing as in Equation \ref{eqn:wce}, where $y$ and $\hat{y}$ are respectively the groundtruth and predicted labels for the operators, $\sigma$ is a sigmoid function, and $w$ is a weight vector for class balancing.
\begin{equation}
L = -1\Big(y\log\big(\sigma(\hat{y})\big)w + (1-y)\log\big(1-\sigma(\hat{y})\big)\Big)
\label{eqn:wce}
\end{equation}

In \textit{weak-} and \textit{self-supervised} learning, the model adopts a Siamese-style approach, contrasting negative or dissimilar pairs and pulling positive or similar pairs closer using the margin-based loss in Equation \ref{eqn:closs}, where $y$ is 0 for negative pairs and 1 for positive pairs, $d$ is the Euclidean distance between two pairs, and $m$ is the margin of distance tolerance, set to 0.5 in our experiment.
\begin{equation}
L = d^2y + \max(m-d,0)^2(1-y)
\label{eqn:closs}
\end{equation}

In the weak supervision scheme, we leverage the knowledge that tools with different track IDs have different operator categories, helping segregate positive and negative pairs.
For the self-supervised setting, we group pairs based on the tool's \textit{direction assumption}: in a frame where there are multiple tools, they are considered opposing pairs and their positive pairs are generated by a soft augmentation (e.g. scale, perspective, slight rotation $\theta{<}10^{\circ}$, mixed). 
Conversely, in frames with a single tool, negative pairs are generated by rotating the tool to a different direction (usually $80^{\circ}\leq\theta\leq100^{\circ}$ and $170^{\circ}\leq\theta\leq190^{\circ}$), and positive pairs are formed through soft augmentation.

\subsection{Identity Association using Multi-Perspective Matching}
The role of the identity association module is to link tool detections across frames, creating coherent tracks for surgical tools. It is a critical step in understanding how these tools move and interact during procedures.
To tackle this challenge effectively, we introduce a novel approach called \textit{Harmonizing Bipartite Graph Matching} (HBGM). This method is designed to handle the diverse perspectives of surgical tool trajectories without conflicts, ensuring that every tool is tracked accurately throughout the surgery.
HBGM builds upon the Bipartite graph matching (BGM) algorithm \citep{kuhn1955hungarian}, which is a powerful tool in its own right. What sets HBGM apart is its ability to manage multiple track states for each tool instance.
These states include:
\begin{enumerate}
    \item {\it New:} Assigned at track initialization.
    \item {\it Active:} Updated when a detection matches a track.
    \item {\it Lost:} Assigned to an active track with no matching detection at current time. The ``Lost" state has a 5-second time-threshold within which the track can be re-activated for all perspectives.    
    \item {\it Out of Camera View (OOCV):} 
    Assigned if a track remains ``Lost" for more than the time-threshold. This state marks the end of a visibility trajectory. An ``OOCV" track can only be re-activated for intraoperative and intracorporeal perspectives.    
    \item {\it Out of Body (OOB):} 
    Assigned if a tool track is judged to have left the abdomen. This is determined when a track fails a class-direction consistency check (described in Section \ref{sec:cdc-check}). The OOB state marks the end of an intracorporeal trajectory. Such track can only be re-activated for intraoperative perspective.    
    \item {\it Removed:}  Assigned if a track is initialized erroneously (e.g., after the ``New" state, it never progresses to ``Active" for a prolonged time) or not reassigned for a long period. Removed tracks are excluded from history against a potential match.
\end{enumerate}
The stages of HBGM are illustrated in Figure \ref{fig:architecture}(c) and discussed as follows:

\subsubsection{Pairwise Matching}
To match tools across frames, we use pairwise distances between predicted motion directions ($\theta$) and detection boxes ($B$), while ensuring class ($C$) consistency. 
For each predicted tool $p_i^t = \{B_i^t, C_i^t, \theta_i^t\}$ in frame $f_t$, and each active track $q_j^{t-1} = \{B_j^{t-1}, C_j^{t-1}, \theta_j^{t-1},~ \text{IDs}\}$ in the history trajectories up to frame $f_{t-1}$, the distance metric $Dist_{i,j}$ is defined as:
\begin{multline}
    \label{eqn:cost}
    \text{Dist}_{i,j} = \alpha \cdot \text{Dist}_{\text{spatial}}(B_i^t, B_j^{t+1}) ~ + \\ \beta \cdot \text{Dist}_{\text{direction}}(\theta_i^t,  \theta_j^{t+1}) ~ + \\ \gamma \cdot \text{Dist}_{\text{class}}(C_i^t, C_j^{t+1}),
\end{multline}

where
\begin{equation}
\text{Dist}_{\text{spatial}}(B_i^t, B_j^{t+1}) = 1 - \frac{|B_i^t \cap B_j^{t+1}|}{|B_i^t \cup B_j^{t+1}|},
\end{equation}
is the Intersection over Union (IoU) similarity between the bounding boxes $B_i^t$ and $B_j^{t+1}$. Alternatively, $Dist_{spatial}(B_i^t, B_j^{t+1})$ can be computed as the Euclidean distance between bounding box centers:
$ \sqrt{ (X_i^t - X_j^{t+1})^2 + (Y_i^t - Y_j^{t+1})^2 }$,

\begin{equation}
\text{Dist}_{\text{direction}}(\theta_i^t, \theta_j^{t+1}) = 1 - \frac{\theta_i^t \cdot \theta_j^{t+1}}{\|\theta_i^t\|\cdot \|\theta_j^{t+1}\|},
\end{equation}
is the angular difference between motion directions, and

\begin{equation}
\text{Dist}_{\text{class}}(C_i^t, C_j^{t+1}) = \begin{cases} 
1-{S} & \text{if } C_i^t = C_j^{t+1} \\
\tau & \text{if } C_i^t \neq C_j^{t+1},
\end{cases}
\end{equation}
ensures that tools of different classes are not matched ($S$ is the class probability score of the tool and $\tau$ is set above matching threshold). Here, $\alpha$, $\beta$, and $\gamma$ are weights that balance the influence of spatial, motion, and class distances, respectively.
A bipartite matching algorithm is then used to minimize $\text{Dist}_{i,j}$ and establish correspondences. Optionally, a second bipartite association is performed for low-confidence detections, following the method proposed by \citet{zhang2022bytetrack}.

\subsubsection{Track Harmonization}
The magic of HBGM happens during the harmonization stage, where it skillfully handles track ID assignment ensuring consistent tool tracking across multiple perspectives — namely intraoperative, intracorporeal, and visibility. 
The harmonization process relies on a structured approach involving (a) class-direction consistency checks, (b) hierarchical ID assignment, and (c) referential integrity to effectively manage tool trajectories under different states and handle potential conflicts in track identification and state updates.\\

{\noindent\bf A. Class-Direction Consistency Check:}
\label{sec:cdc-check}
HBGM validates that a track maintains a joint matching of class ($C$) and direction ($\theta$) features across time to remain active. This check is specifically for managing the intracorporeal trajectory.
A change in the class feature (or class ID) while the direction feature remains unchanged indicates that a new tool $j$ has been introduced through the same trocar as previous tool $i$, marking the end of the tool $i$ intracorporeal trajectory and the start of a new one:
\begin{equation}
    \text{If } \theta_i^t = \theta_j^{t-1} \text{ and } C_i^t \neq C_i^{t-1}, \text{ then } \text{State}_j^{t-1} = \text{OOB}.
\end{equation}
This approach enforces a one-to-one correspondence between track direction and class features for more stable and accurate tracking.\\

{\noindent\bf B. Hierarchical ID Initialization and State Updates:}
HBGM employs a hierarchical approach to manage new ID assignments and state updates effectively across perspectives.
At the topmost level is the intraoperative  trajectory ($p=3$) followed by the intracorporeal trajectory ($p=2$), and lastly the visibility trajectory ($p=1$).

\begin{enumerate}
    \item {\it ID Initialization:} A new ID in a higher perspective ($p=3$) triggers new IDs in all lower perspectives ($p=2$, $p=1$). Similarly, new IDs initialized in $p=2$ are also initialized in $p=1$.
        
    \item {\it State Updates:} State changes in lower perspectives are propagated in higher ones. For example,``Lost" in $p=1$ implies ``Lost" in $p=2$ and $p=3$. ``OOCV" in $p=2$ implies ``OOCV" in $p=3$.
    Hierarchical state update is excluded for ``New" state, which is treated as ID initialization.   
\end{enumerate}

{\noindent\bf C. Referential Integrity:}
Once a tool's track is established in one perspective, every changes in its state is concurrently reflected across all other perspectives.
HBGM improves consistent track assignment for unmatched detections and untracked tracks by cross-referencing of historical data across perspectives. This minimizes ID switches and facilitates track recovery after occlusion, OOCV or OOB.
Also, persistent trajectory patterns in $p=3$ is relied on to dynamically adjust thresholds for the inactive tracks, aiding in robust recovery.

\subsubsection{Track Recovery}
Before initializing new tracks, the algorithm attempts to recover unmatched detections at time $t$ by pairing them with inactive tracks as at time ${t-1}$ using only directional ($\theta$) and categorical ($C$) features. 
For each unmatched detection $d'_i \in \mathbf{D}_{\text{unmatched}}^{t}$ and inactive track $d_j \in \mathbf{D}_{\text{inactive}}^{t-1}$, a recovery score $S_{ij} = \alpha \cdot \text{Dist}_\text{direction}(\theta_i', \theta_j) + \beta \cdot \mathbb{1}(C_i' \neq C_j)$ is computed, where $\text{Dist}_{direction}(\theta_i', \theta_j)$ measures directional similarity and $\mathbb{1}(C_i' \neq C_j)$ indicates categorical mismatch. The unmatched detection $d'_i$ is assigned to the inactive track $d_j$ with the lowest score, provided it falls below a predefined threshold $\tau_{\text{rec}}$. If no valid match is found, a new track ID is initialized. This approach leverages feature consistency to ensure seamless recovery and reduces the chance of ID switching or track loss.

\subsubsection{Versatility}  
Optionally, HBGM can seamlessly integrate other re-identification features, such as the Kalman filter \citep{kalman1960new}, appearance features \citep{zhang2021fairmot}, camera motion compensation \citep{aharon2022bot}, and similarity features \citep{wang2023smiletrack}.

The Kalman filter \citep{kalman1960new} enhances the spatial distance metric term ($\text{Dist}_{\text{spatial}}$) in Eqn.~\eqref{eqn:cost} by predicting the future positions of track bounding boxes based on their previous states. The Euclidean distance between these predicted positions and the current bounding box centers is then calculated, yielding a more robust spatial cost component.
For appearance and other re-ID features, a corresponding distance cost is computed similarly to the direction features ($\text{Dist}_{\text{direction}}$). Each re-ID feature contributes its distance term, which is weighted and added to the overall distance metric $\text{Dist}_{i,j}$ in Eqn. \eqref{eqn:cost} before applying bipartite matching for track assignment.

Overall, our approach ensures accurate track ID reassignment, mitigates identity fragmentation and switches, and provides fine-grained track details and a rich trajectory history.

\subsection{Baseline Method}
To demonstrate the efficacy of our sophisticated modeling, we design a\textit{ knowledge-based (KB) baseline}, integrating predefined surgical states and priors into existing trackers.
In the context of laparoscopic cholecystectomy, approximately 75\% of the utilized tools in a procedure exhibit unique categories and a known instance count, applicable solely within the intraoperative trajectory perspective. The KB baseline is a straightforward, non-trainable algorithm that incorporates predefined constraints:
\begin{enumerate}
    \item {\bf Inter-Class Switch Constraint:} This utilizes the tool class features to disallow identity switch across tool categories with inter-class bounding box overlap, mostly observed in crowded scenes and tool occlusion scenarios.
    \item {\bf Maximum Instance Constraint:} To mitigate identity fragmentation, a new track assignment is allowed only if the assigned trajectories per tool category remain within the allowable maximum track instances otherwise the matching threshold is incrementally lowered until the candidate tool is matched to a probable existing trajectory. 
    Attempts to extend the application of this constraint beyond intraoperative tracking, with non-deterministic instance count, yield no meaningful results.
\end{enumerate}

Aside from the KB baseline, we benchmark over 20 SOTA Deep Learning methodologies on CholecTrack20 dataset for surgical tool detection, re-identification, and tracking.
We adapt some of the SOTA tracking models for MCMOT to suit the surgical context of our study.

\section{Experiments}
\label{sec:exps}

\subsection{Implementation Details}
The proposed model is implemented in PyTorch; implementation details and hyperparameters are tabulated in Table \ref{tab:exp}.
Depending on code availability, baseline models are outsourced from public repositories or re-implemented.
We follow the official splits of the CholecTrack20 \citep{Nwoye2023CholecTrack20} dataset to set 10, 2, and 8 videos for training, validation, and testing respectively.

\subsection{Evaluation Metrics}
We assess the quality of the direction embeddings for track re-identification through a consistency accuracy metric: here, a predicted tool's direction is correct if the pairwise difference between its embeddings at time $t$ and $t-1$ is less than a predefined threshold (usually $\theta=0.5$). We utilize extended time intervals $t{-}k$, comparing embeddings consistency up till the past k frames, $k\in [1,5,25,\text{start}]$.

We follow the recommended protocol and metrics for CholecTrack20 \citep{Nwoye2023CholecTrack20} tool tracking evaluation.
Here, we evaluate different aspects of the tracking performance using the standard metrics of higher-order tracking accuracy (HOTA) \citep{luiten2021hota} (including its constituents: Localization Accuracy, LocA, Detection Accuracy, DetA, and Association Accuracy, AssA), the CLEAR metrics \citep{bernardin2008evaluating} of multi-object tracking accuracy (MOTA), multi-object tracking precision (MOTP), mostly tracked (MT), mostly lost (ML), partially tracked (PT), identity switch (IDSW), track fragmentation (Frag). We also assess the identity F1 score (IDF1) \citep{ristani2016performance}, counting metrics (number of detections \#Dets and identities \#IDs), and tracking speed in frame per second (FPS).

\begin{table}[!t]
    \centering
    \caption{Methods implementation details.}
    \label{tab:exp}    
    \resizebox{\columnwidth}{!}{%
    \begin{tabular}{|ll|ll|}
        \hline
        Pre-training & MOT20, CH & Learning rate & 3e-4 \\
        Pre-processing & Crop, Resize & Batch size & 32 \\
        Post-processing & NMS & Epochs & 132 \\
        Data augmentation & Soft, Negative & Optimizer & Adam \\
        Detector input shape & 542 x 655 & Weight decay & 1e-5 \\
        Re-ID input shape & 224 x 224 & Model params & 123M  \\
        Train/val/test splits & 10/2/8 videos & Lr decay schedule & Plateaux \\
        
        Detector Loss func. & BCE & GPU resources & 4 $\times$ V100 \\
        FSL Re-ID Loss & BCE  & Exp. training time & 48 h \\
        WSL Re-ID Loss & Cosine CL & Baseline training time & 139 h \\
        SSL Re-ID Loss & Contrastive & Ablation training time & 120 h \\
        \hline
    \end{tabular}
    }
\end{table}

\section{Results and Discussion}
\label{sec:results}
First, our base detector, YOLOv7 \citep{wang2023yolov7}, yields $80.6\%~AP_{0.5}$ and $56.1\%~AP_{0.5:0.95}$ for tool detection at an inference speed of 20.6 FPS, demonstrating its effectiveness as a detector for our tracking model. Additional results on the tool detection can be found in \citet{Nwoye2023CholecTrack20}.

\begin{table*}[!th]
    \centering
    \caption{Surgeon Operator Prediction based on Direction Feature Embeddings for Tool Track Re-identification - Results showing the mean and per-class Average Precision (\% AP) and embedding temporal consistency (\% Accuracy) from time ${t-k}$ to $t$. (Grasper Only).}
    \label{tab:results-operators}
    \setlength{\tabcolsep}{4pt}
    \resizebox{0.9\linewidth}{!}{%
    \begin{tabular}{@{}lcccccccclcccclcccc@{}}
        \toprule         
        \multirow{2}{*}{Model} &
        \multicolumn{8}{c}{Full Supervision} & \phantom{abc} &
        \multicolumn{4}{c}{Weak Supervision} & \phantom{abc} &
        \multicolumn{4}{c}{Self Supervision}\\         
        \cmidrule{2-9} \cmidrule{11-14} \cmidrule{16-19}
        & MSLH & ASRH & MSRH & mAP & $T^t_{t-1}$ & $T^t_{t-5}$ & $T^t_{t-25}$ & $T^t_{t_0}$ &&  $T^t_{t-1}$ & $T^t_{t-5}$ & $T^t_{t-25}$ & $T^t_{t_0}$ && $T^t_{t-1}$ & $T^t_{t-5}$ & $T^t_{t-25}$ & $T^t_{t_0}$ \\        
        \midrule \midrule
        Siamese Baseline \citeb{he2018twofold} & 92.3 & 81.5 & 11.2 & 61.7 & 80.0 & 78.1 & 74.7 & 68.1 && \bf 64.6 & \bf 62.1 & 58.2 & \bf 51.1 && 62.1 & 61.0 & 58.6 & 47.8 \\
        ViT \citeb{dosovitskiy2020image} & 92.1 & 77.7 & 10.2 & 60.0 & 78.7 & 76.7 & 73.2 & 67.3 && 63.0 & 58.7 & 53.4 & 50.3 && 64.8 & 62.1 & 57.7 & 51.0 \\
        CrossViT \citeb{chen2021crossvit} & 91.7 & 79.6 & 9.7 & 60.3 & 78.5 & 76.8 & 73.9 & 69.8 && 61.7 & 60.0 & 57.5 & 50.3 && 61.8 & 60.7 & 58.2 & 47.4\\
        ResNet-18 \citeb{he2016deep} & 95.5 & 87.1 & 36.6 & 73.0 & 81.3 & 79.2 & 76.3 & 72.8 && 62.1 & 60.8 & 57.7 & 46.5 && 87.6 & 86.7 & 85.4 & 83.5 \\
        Efficientnet-B0 \citeb{tan2019efficientnet} & \textbf{97.5 }& 92.1 & 32.5 & 74.0 & 86.0 & 84.5 & 82.7 & 81.3 && \textbf{64.6} & 61.5 & 55.3 & 42.8 && 89.7 & 88.7 & 87.3 & 85.1 \\
        SMILETrack \citeb{wang2023smiletrack} & 89.8 & 75.7 & 13.6 & 59.7 & 78.5 & 76.9 & 73.7 & 62.3 && 62.0 & 60.0 & 54.5 & 50.6 && 61.7 & 60.4 & 56.7 & 46.1 \\
        Proposed Estimator & \bf 97.5 & \bf 93.0 & \bf 53.0 & \bf 81.2 & \bf 91.0 & \bf 89.0 & \bf 88.6 & \bf 87.4 && 63.6 & 61.8 & \bf 58.3 & 47.5 && \bf 91.6 & \bf 90.7 & \bf 89.9 & \bf 88.4 \\
        \bottomrule
    \end{tabular}
    }
\end{table*}

\subsection{Results of Surgeon Operator Prediction}
We measure the quality of the direction features used for track re-identification in this task.
The proposed Estimator, built on the EfficientNet-B0 \citep{tan2019efficientnet} backbone, demonstrates remarkable performance in surgical tool re-identification across different supervision settings as evidenced by the results presented in Table \ref{tab:results-operators}. Leveraging EfficientNet-B0's known efficiency and speed, it outperforms several strong baselines: Siamese baseline, ResNet \citep{he2016deep}, ViT \citep{dosovitskiy2020image}, CrossViT \citep{chen2021crossvit}, etc., in capturing essential features for tool re-identification and achieving the highest mean Average Precision (mAP) of 81.2\% under supervised setting. 
The inclusion of an attention head enhances its ability to learn direction features, which is crucial for distinguishing between similar tool instances, thereby outperforming the baseline EfficientNet-B0 (+ 4.2\% mAP).
Category-wise, the main surgeon's right hand (MSRH), which is the busiest and handles most of the tools, exhibits the greatest detection difficulty with a 53\% AP.

In the self-supervised setting, the proposed Estimator showcases its versatility by consistently achieving high re-identification accuracy ($\geq 88\%)$ over various time intervals, including longer time differences up to the start of the video (e.g., $T^t_{t-25}$ and $T^t_{t_0}$). This demonstrates that the learned direction-aware features effectively maintain consistency, even when dealing with challenging long-term tracking scenarios; 
thanks to the innovative image preprocessing technique of ``\textit{image slicing and padding}'' that effectively addresses the visibility issue of tool shafts in images, enhancing the model's ability to capture directional information.
The direction features align closely with the tool operator's hand direction from the trocar port, emphasizing their relevance as robust re-identification features for surgical tool tracking. This analysis focus on grasper because it is the only tool with multiple instance when considering intraoperative trajectory.
We, however, observe a slightly inferior performance when models are weakly-supervised on track ID labels, which is the same supervisory signal for conventional appearance re-ID models \citep{zhang2022bytetrack,aharon2022bot,wang2023smiletrack}.

\subsection{Ablation Study on Re-ID Features}

The ablation study in Table \ref{tab:ablation-features} explores the impact of track re-identification (Re-ID) features on multi-object tracking, categorizing them into {parametric and non-parametric groups}.
The non-parametric features are bounding box location (IoU), low-confidence detection (Byte) \citep{zhang2022bytetrack}, motion prediction using Kalman filter (KF) \citep{kalman1960new}, camera-motion compensation (CMC) using BoT-SORT-ReID \citep{aharon2022bot}, and multi-class (MC) identity labels.
The parametric features are appearance features (AF) \citep{zhang2021fairmot}, similarity features (SF) \citep{wang2023smiletrack}, and the proposed direction features (DF) - Using the SSL paradigm.

\begin{table}[!tb]
    \centering
    \caption{Ablation study on track re-identification features. Using the intraoperative trajectory perspective @ 25 FPS.}
    \label{tab:ablation-features}
    \setlength{\tabcolsep}{4pt}
    \resizebox{0.999\columnwidth}{!}{%
    \begin{tabular}{@{}lcccccccclcccc@{}}
        \toprule
        &\multicolumn{8}{c}{Re-ID features} & \phantom{abc} & \multicolumn{4}{c}{Tracking Results} \\
        \cmidrule{2-9} \cmidrule{11-14}
         & IOU & BYTE & KF & CMC & MC & AF & SF & DF && HOTA$\uparrow$ & MOTA$\uparrow$ & IDF1$\uparrow$ & \\ 
        \midrule        
        \midrule        
        \multirow{7}{*}{\rot{90}{Non-parametric}}
        &\used&\null&\null&\null&\null&\null&\null&\null&& 14.8 & 57.3 & 8.7 & \\ 
        &\used&\used&\null&\null&\null&\null&\null&\null&& 14.7 & 59.2 & 8.8 & \\ 
        &\used&\null&\used&\null&\null&\null&\null&\null&& 13.9 & \bf 62.2 & 7.7 & \\ 
        &\used&\null&\null&\used&\null&\null&\null&\null&& 13.5 & 60.7 & 7.3 & \\ 
        &\used&\null&\null&\null&\used&\null&\null&\null&& 16.6 & 58.7 & 9.6 & \\ 
        &\used&\used&\null&\null&\used&\null&\null&\null&& \bf 16.8 & 57.2 & \bf 9.8 & \\ 
        &\used&\used&\used&\used&\used&\null&\null&\null&& 13.9 & 60.5 & 7.5 & \\ 
        \midrule       
        \multirow{7}{*}{\rot{90}{Parametric}} 
        &\null&\null&\null&\null&\null&\used&\null&\null&& 19.1 & 51.3 & 12.1 & \\ 
        &\null&\null&\null&\null&\null&\null&\used&\null&& 25.5 & 40.9 & 19.4 & \\ 
        &\null&\null&\null&\null&\null&\null&\null&\used&& 59.5 & 72.5 & 65.4 & \\ 
        &\null&\null&\null&\null&\null&\used&\used&\null&& 32.4 & 73.3 & 31.9 & \\ 
        &\null&\null&\null&\null&\null&\used&\null&\used&& \bf 62.6 & \bf 78.8 & \bf 71.4 & \\ 
        &\null&\null&\null&\null&\null&\null&\used&\used&& 62.2 & 77.7 & 70.9 & \\ 
        &\null&\null&\null&\null&\null&\used&\used&\used&& 62.0 & 78.2 & 70.3 & \\ 
        \midrule
        \multirow{11}{*}{\rot{90}{Joint}}        
        &\used&\null&\null&\null&\null&\used&\null&\null&& 22.8 & 64.8 & 16.3 & \\ 
        &\used&\null&\null&\null&\null&\null&\used&\null&& 29.1 & 49.7 & 25.0 & \\ 
        &\used&\null&\null&\null&\null&\null&\null&\used&& \bf 61.3 & 78.5 & \bf 67.8 & \\ 
        &\used&\null&\null&\null&\used&\null&\null&\used&& 60.2 & \bf 78.8 & 66.8 & \\ 
        &\used&\used&\null&\null&\used&\null&\null&\used&& 57.9 & 75.9 & 63.6 & \\ 
        \cmidrule{2-14}        
        &\used&\null&\null&\null&\null&\used&\used&\null&& 25.2 & 66.7 & 20.7 & \\ 
        &\used&\null&\null&\null&\null&\used&\null&\used&& 60.7 & \bf 78.9 & 67.6 & \\ 
        &\used&\null&\null&\null&\null&\null&\used&\used&& 60.6 & 77.8 & 68.0 & \\ 
        &\used&\null&\null&\null&\null&\used&\used&\used&& \bf 61.5 & 78.5 & \bf 69.3 & \\ 
        &\used&\used&\null&\null&\used&\used&\used&\used&& 57.6 & 77.3 & 64.2 & \\ 
        &\used&\used&\used&\used&\used&\used&\used&\used&& 55.5 & 65.4 & 62.1 & \\ 
        \bottomrule
    \end{tabular}
    }
\end{table}

In the non-parametric settings, IoU, BYTE, and KF show incremental contributions, with IoU + BYTE + MC achieving the highest HOTA (16.8\%). Parametric features like AF, SF, and DF exhibit significant individual capabilities, with DF standing out at 59.5\% HOTA. Combinations of AF + SF (32.4\%) and AF + DF (62.6\%) highlight synergies between appearance and directional features. Jointly, IoU + MC + DF excels (60.2\%), and further additions (BYTE, AF) refine results, culminating in the optimal IoU + AF + SF + DF (61.5\%) configuration. This underscores the nuanced interplay of Re-ID features, with DF proving effective independently, synergistically with AF and SF, and being judiciously combinable for enhanced tracking performance. 
The study emphasizes the versatility of DF, showing its competence with basic geometric features and in comprehensive multi-feature scenarios, providing insights for real-time applications.
Incorporating DF, as shown through the ablation experiments in Table \ref{tab:ablation-features}, usually leads to a more significant improvement compared to incorporating appearance-based features, suggesting that the DF capture more than just visual and spatial cues.

\subsection{Ablation Study on Track Association Algorithm}

\begin{table}[!tb]
    \centering
    \caption{Ablation study on linear association algorithms and approach for combining multiple re-ID costs.} 
    \label{tab:ablation-assoc}
    \setlength{\tabcolsep}{4pt}    
    \resizebox{0.999\columnwidth}{!}{%
    \begin{tabular}{cclcccccclccc}
        \toprule
        \multicolumn{2}{c}{Algorithm} & \phantom{abd}&\multicolumn{6}{c}{Re-ID Costs Ensemble} & \phantom{abc} & \multicolumn{3}{c}{Results}\\
        \cmidrule{1-2} \cmidrule{4-9} \cmidrule{11-13}
        
        BM & HBM && Min & Avg & w.Avg & Vote & w.Vote & w.A.V && HOTA$\uparrow$ & MOTA$\uparrow$ & IDF1$\uparrow$ \\
        \midrule  
        \midrule      
        \used &&&\used & & & & & & & 12.2 & 23.1 & 18.3 \\ 
        &\used &&\used & & & & & & & 29.3 & 58.5 & 26.8 \\ 
        &\used & &&\used & & & & & & 37.4 & 79.9 & 39.8 \\ 
        &\used & && &\used & & & & & 36.7 & 74.9 & 33.2 \\ 
        &\used & && & &\used & & & & 56.4 & \bf 80.1 & 60.0 \\ 
        &\used & && & & &\used & & & 58.0 & 74.4 & 63.1 \\ 
        &\used & && & & & &\used & & \bf 61.5 & 78.5 & \bf 69.3 \\ 
        \bottomrule
    \end{tabular}
    }
\end{table}

\begin{table*}[!t]
    \centering
    \caption{Multi-Perspective Multi-Tool Tracking Results @ 25 FPS.}
    \label{tab:benchmark-trackers}
    \setlength{\tabcolsep}{3pt}
    \resizebox{0.94\linewidth}{!}{%
    \begin{tabular}{@{}lcccclccccclccclcclr@{}}
        \toprule
        Model & 
        \multicolumn{4}{c}{HOTA Metrics} &\phantom{abc}&
        \multicolumn{5}{c}{CLEAR Metrics} &\phantom{abc}&
        \multicolumn{3}{c}{Identity Metrics} &\phantom{abc}&
        \multicolumn{2}{c}{Count Metrics} &\phantom{abc}&
        \multicolumn{1}{c}{Speed} \\        
        \cmidrule{2-5}
        \cmidrule{7-11}
        \cmidrule{13-15}
        \cmidrule{17-18}
        \cmidrule{20-20}
        Tracker 
        & HOTA$\uparrow$ 
        & DetA$\uparrow$
        & LocA$\uparrow$ 
        & AssA$\uparrow$
        &        
        & MOTA$\uparrow$
        & MOTP$\uparrow$
        & MT$\uparrow$
        & PT$\downarrow$
        & ML$\downarrow$
        &
        & IDF1$\uparrow$ 
        & IDSW$\downarrow$
        & Frag$\downarrow$
        &
        & \#Dets
        & \#IDs
        &
        & FPS$\uparrow$ \\
        \midrule \midrule        
        &\multicolumn{18}{c}{\normalsize\bf\it Intraoperative Trajectory (Groundtruth counts: \#Dets = 29994, \#IDs = 70)}\\
        \midrule
        OCSORT \citeb{maggiolino2023deep} & 14.6 & 52.7 & \bf 86.7 & 4.1 && 49.2 & \bf 85.0 & 24 & 32 & 14 && 9.5 & 2921 & 2731 && 21936 & 3336 && 10.2\\        
        FairMOT \citeb{zhang2021fairmot}& 5.8 & 25.8 & 75.9 & 1.3 && 5.0 & 73.9 & 3 & 24 & 43 && 4.3 & 4227 & 1924 && 15252 & 4456 && 14.2 \\        
        TransTrack \citeb{sun2020transtrack} & 7.4 & 31.5 & 84.4 & 1.7 && 4.2 & 82.9 & 9 & 36 & 25 && 4.2 & 4757 & \bf 1899 && 21640 & 4079 && 6.7\\        
        ByteTrack \citeb{zhang2022bytetrack} & 15.8 & 70.6 & 85.7 & 3.6 && 67.0 & 84.0 & 54 & 12 & 2 && 9.5 & 4648 & 2429 && 28440 & 5383 && 16.4 \\        
        Bot-SORT \citeb{aharon2022bot} & 17.4  & 70.7 & 85.4 & 4.4 && 69.6 & 83.7 & \bf 58 & 11 & \bf 1 && 10.2 & 3907 & 2376 && 29302 & 4501 && 8.7 \\        
        SMILETrack \citeb{wang2023smiletrack} & 15.9 & 71.0 & 85.5 & 3.7 && 66.4 & 83.8 & 55 & 13 & 2 && 9.2 & 4968 & 2369 && 28821 & 5761 && 11.2 \\ 
        \midrule
        ByteTrack (KB) & 36.5 & 70.1 & 85.4 & 19.1 && 74.5 & 83.7 & \bf 58 & 10 & 2 && 33.3 & 2046 & 2406 && \bf 29504 & 427 && 16.3 \\        
        Bot-SORT (KB) & 37.4  & 67.8 & 85.4 & 20.7 && 73.5 & 83.6 & 51 & 12 & 7 && 35.5 & 1638 & 2199 && 27605 & 315 && 8.6 \\        
        SMILETrack (KB) & 37.5 & 65.7 & 85.6 & 21.6 && 71.4 & 83.8 & 49 & 13 & 8 && 35.7 & 1444 & 2021 && 26517 & 266 && 11.0 \\
        \midrule
        SurgiTrack (FSL) & \bf 67.3 & 70.8 & \bf 86.7 & \bf 64.1 && 76.0 & \bf 85.0 & 48 & 20 & 2 && \bf 81.7 & 1891 & 2489 && 27499 & \bf 72 && 15.3 \\     
        SurgiTrack (WSL) & 56.5 & \bf 71.7 & 86.5 & 44.7 && \bf 78.9 & 84.8 & 54 & 14 & 2 && 60.3 & \bf 1352 & 2368 && 28120 & 79 && 15.3 \\      
        SurgiTrack (SSL) & 60.2 & \bf 71.7 & 86.6 & 50.6 && 78.8 & 84.8 & 54 & 14 & 2 && 66.8 & 1373 & 2364 && 28102 & 85 && 15.3 \\
        \midrule \midrule        
        &\multicolumn{18}{c}{\normalsize\bf\it Intracorporeal Trajectory (Groundtruth counts: \#Dets = 29994, \#IDs = 247)}\\
        \midrule
        OCSORT \citeb{maggiolino2023deep} & 23.7 & 51.4 & 86.5 & 11.0 && 47.1 & 84.8 & 115 & 87 & 45 && 18.1 & 2953 & 2796 && 21797 & 3526 && 10.2 \\
        FairMOT \citeb{zhang2021fairmot}& 7.5  & 19.7 & 76.1 & 2.9 && 5.4 & 74.0 & 19 & 60 & 168 && 6.0 & 2890 & 1496 && 11287 & 3962 && 14.2 \\
        TransTrack \citeb{sun2020transtrack} & 13.1  & 31.5 & 84.4 & 5.5 && 4.6 & 82.9 & 80 & 79 & 88 && 8.7 & 4648 & \bf 1791 && 21640 & 4079 && 6.7 \\
        ByteTrack \citeb{zhang2022bytetrack} & 24.7  & 70.6 & 85.7 & 8.7 && 67.4 & 84.0 & 176 & 48 & 23 && 16.9 & 4515 & 2290 && 28440 & 5383 && 16.4 \\
        Bot-SORT \citeb{aharon2022bot} & 27.0 & 70.7 & 85.4 & 10.4 && 70.0 & 83.7 & \bf 188 & 38 & \bf 21 && 18.9 & 3771 & 2238 && \bf 29300 & 4501 && 8.7 \\
        SMILETrack \citeb{wang2023smiletrack} & 24.9  & 66.7 & 85.5 & 8.9 && 66.7 & 83.8 & 186 & 39 & 22 && 16.9 & 4868 & 2232 && 28820 & 5779 && 11.2 \\
        \midrule
        SurgiTrack (FSL) & \bf 55.2 & 70.8 & \bf 86.6 & \bf 43.2 && 74.8 & \bf 84.9 & 170 & 55 & 22 && \bf 61.7 & 2257 & 2350 && 27499 & \bf 1092 && 15.3 \\
        SurgiTrack (WSL) & 34.6 & \bf 71.7 & 86.5 & 16.8 && 75.7 & 84.8 & 184 & 41 & 22 && 28.9 & 2300 & 2229 && 28120 & 2257 && 15.3 \\
        SurgiTrack (SSL) & 39.4 & \bf 71.7 & 86.5 & 21.8 && \bf 77.5 & 84.8 & 184 & 42 & \bf 21 && 36.2 & \bf 1761 & 2225 && 28102 & 1480 && 15.3 \\
        \midrule \midrule        
        &\multicolumn{18}{c}{\normalsize\bf\it Visibility Trajectory (Groundtruth counts: \#Dets = 29994, \#IDs = 916)}\\
        \midrule
        SORT \citeb{bewley2016simple}& 17.4 & 39.5 & 85.2 & 7.8 && 21.4 & 83.3 & 139 & 399 & 378 && 13.4 & 6619 & 2138 && 16595 & 8844 && 19.5 \\        
        OCSORT \citeb{maggiolino2023deep} & 37.0 & 52.6 & 86.5 & 26.2 && 50.2 & 84.8 & 300 & 371 & 245 && 35.9 & 2317 & 2260 && 22197 & 3587 && 10.2 \\
        FairMOT \citeb{zhang2021fairmot}& 15.3 & 25.0 & 75.8 & 9.5 && 7.1 & 73.7 & 58 & 218 & 640 && 14.4 & 3140 & 1574 && 15338 & 4875 && 14.2 \\
        TransTrack \citeb{sun2020transtrack} & 19.2 & 31.6 & 84.4 & 11.8 && 5.8 & 82.9 & 224 & 280 & 412 && 16.1 & 4273 & \bf 1403 && 21640 & 4079 && 6.7 \\
        ByteTrack \citeb{zhang2022bytetrack} & 41.5 & 70.7 & 85.7 & 24.8 && 69.3 & 84.0 & 591 & 217 & 108 && 36.8 & 3930 & 1704 && 28440 & 5383 && 16.4 \\
        Bot-SORT \citeb{aharon2022bot} & 44.7 & 70.8 & 85.5 & 28.7 && 72.0 & 83.7 & \bf 638 & 184 & \bf 94 && 41.4 & 3183 & 1638 && \bf 29300 & 4505 && 8.7 \\
        SMILETrack \citeb{wang2023smiletrack} & 41.3 & 71.0 & 85.6 & 24.4 && 68.9 & 83.8 & 619 & 192 & 105 && 36.5 & 4227 & 1641 && 28821 & 5752 && 11.2 \\
        \midrule
        SurgiTrack (FSL) & 61.6 & 70.9 & \bf 86.6 & 53.9 && 75.4 & \bf 84.9 & 551 & 244 & 121 && 68.9 & 2095 & 1778 && 27499 & \bf 1814 && 15.3 \\
        SurgiTrack (WSL) & 58.4 & \bf 71.7 & 86.5 & 47.8 && 76.8 & 84.8 & 601 & 201 & 114 && 62.3 & 1980 & 1650 && 28120 & 2678 && 15.3 \\
        SurgiTrack (SSL) & \bf 62.8 & \bf 71.7 & 86.5 & \bf 55.3 && \bf 78.0 & 84.8 & 598 & 202 & 116 && \bf 69.4 & \bf 1546 & 1648 && 28102 & 2153 && 15.3 \\        
        \bottomrule
    \end{tabular}
    }
\end{table*}

The comparative analysis between Bipartite Graph Matching (BGM) and Harmonizing Bipartite Graph Matching (HBGM) in Table \ref{tab:ablation-assoc} reveals a substantial enhancement with HBGM, exhibiting a noteworthy increase of +17.1\% HOTA, +35.4\% MOTA, and +8.5\% IDF1. 
Regarding the use of multiple re-ID features,
we explore the location, appearance, similarity, and direction cost matrices and examine several possible ensemble approaches.
Contrary to the commonly used minimum (Min) method, which tends to bias predictions toward the least cost matrix, ensemble methods such as averaging (Avg) and voting (Vote) gives higher scores showcasing the importance of consensus.
The introduction of weighted averages (w.Avg) and weighted voting (w.Vote) results in a more optimal solution, emphasizing the significance of weighting by the individual strengths of the re-ID features for more robust tracking. 
Specifically, averaging minimizes tracking error by reducing variances leading to improved MOTA score. Voting improves identity assignment consensus resulting in higher HOTA and IDF1 scores.
Combining the strengths of weighted averaging and voting (w.A.V) balances and improves the tracking performances.

\subsection{Analysis of Multi-Tool Tracking Results}
We evaluate our method on the CholecTrack20 dataset, and compare it with the state-of-the-art and baseline methods (Table \ref{tab:benchmark-trackers}). 
The existing models, including OCSORT \citep{maggiolino2023deep}, TransTrack \citep{sun2020transtrack}, ByteTrack \citep{zhang2022bytetrack}, Bot-SORT \citep{aharon2022bot}, and SMILETrack \citep{wang2023smiletrack}, exhibit varying degrees of performance across HOTA metrics. 
While these models show good detection and localization capabilities with relatively high DetA, LocA, and MOTP scores, they struggle with tool track identity association, evident in their low AssA and IDF1 scores. 
TransTrack is affected the most with a record low HOTA of 7.4\%. ByteTrack, Bot-SORT, and SMILETrack, although relatively better, still show room for improvement in HOTA scores ranging from 15.7\% to 17.4\%.
This illustrates that, given the similarity of most tools especially the ones from the same category, relying on location and appearance/similarity features is not sufficient for their correct re-identification across time.

The infusion of surgical knowledge base (KB) into ByteTrack, Bot-SORT, and SMILETrack (resulting in KB variants) helps the identity association and brings about significant improvements, particularly in HOTA scores, ranging from 36.5\% to 37.5\%. 
Despite these enhancements, the proposed model, SurgiTrack, surpasses all existing and KB-infused models, achieving an outstanding HOTA score of 67.3\% in the fully supervised setting.
SurgiTrack excels in individual HOTA metrics, with high DetA (70.8\%) and LocA (86.7\%), emphasizing its precision in tool detection and localization. The robust AssA score of 64.1\% underscores SurgiTrack's proficiency in tool association or re-identification, thanks to the tool direction features.

Comparatively, the SurgiTrack variants trained in weakly supervised (WSL) or self-supervised (SSL) manners also demonstrate strong performance, with SSL showing a slightly higher AssA of 50.6\%. Notably, WSL and SSL produce the highest MOTA and lowest ID switches.
Being the more competitive variant, the SSL-based SurgiTrack is preferred in scenarios with limited annotated data. 
Our analysis also cover the counts of number of detections made, number of unique identities assigned, and the tracking speed as illustrated in Table \ref{tab:benchmark-trackers}.
SurgiTrack shows closer to the groundtruth identity counts.
These results demonstrate the effectiveness of our proposed method and underscore its potential as an advanced solution for multi-tool tracking in intraoperative scenarios.

\begin{table*}[!tb]
    \centering
    \caption{Class-wise tracking accuracy and Impact of KB algorithm on state of the art models; [Intraoperative Perspective @ 25 FPS].}
    \label{tab:classwise}    
    \setlength{\tabcolsep}{6pt}
    \resizebox{0.9\linewidth}{!}{%
    \begin{tabular}{clrcccrccccccc}
        \toprule
        \multicolumn{2}{c}{\multirow{2}{*}{Method}}& \multirow{2}{*}{HOTA} & \multirow{2}{*}{MOTA} & \multirow{2}{*}{IDF1} & & \multicolumn{7}{c}{Class-wise HOTA} \\
        \cmidrule{7-14}
        &&&&& & grasper & bipolar & hook & scissors & clipper & irrigator & spec.bag \\
        \midrule\midrule
         \multirow{3}{*}{Without KB}
         & ByteTrack & 15.8 & 67.0 & 9.5 &
            & 13.3 & 18.0 & 10.5 & 16.3 & 18.3 & 5.7 & 17.6 \\
         & Bot-SORT & 17.4 & 69.6 & 10.2 &
            & 14.3 & 19.6 & 11.4 & 16.9 & 19.1 & 6.4 & 18.7 \\
         & SMILEtrack & 15.9 & 66.4 & 9.2 &
            & 13.0 & 18.0 & 10.4 & 17.0 & 18.0 & 5.9 & 18.8 \\         
         \midrule         
         \multirow{3}{*}{With KB}
         & ByteTrack & 36.5 & 74.5 & 33.3 &
            & 29.1 & 24.2 & 41.3 & 14.6 & 22.7 & 12.5 & 24.1 \\
         & Bot-SORT & 37.4 & 73.5 & 35.5 &
            & 33.0 & 28.9 & 39.8 & 16.3 & 27.7 & 14.6 & 19.1 \\
         & SMILEtrack & 37.5 & 71.4 & 35.7 &
            & 31.9 & 30.4 & 40.8 & 17.2 & 29.2 & 14.5 & 24.8 \\        
         \midrule         
         \multirow{3}{*}{Proposed}
         & SurgiTrack (FSL) & \bf 67.3 & 76.0 & \bf 81.7 &
         & \bf 60.8 & \bf 49.3 & \bf 75.6 & \bf 26.6 & \bf 40.9 & \bf 24.9 & \bf 38.1 \\
         & SurgiTrack (WSL) & 56.5 & \bf 78.9 & 60.3 &
         & 36.8 & 47.9 & 75.0 & 25.5 & 39.4 & 24.0 & 37.2 \\
         & SurgiTrack (SSL) & 60.2 & 78.8 & 66.8 &
         & 45.8 & 48.0 & 75.1 & 25.5 & 39.5 & 24.1 & 37.2 \\
         \bottomrule
    \end{tabular}
    }
\end{table*}

\subsection{Analysis of Multi-Perspective Tracking Results}
Looking at the different trajectory perspectives, it is shown in Table \ref{tab:benchmark-trackers} that visibility tracking is the easiest with most of the existing models showcasing their strengths. This is expected because deep learning models mostly rely on visual cues, which are captured by camera in the visibility track scenario. 
The SSL variant of SurgiTrack record a landslide top performance scores of 62.8\% HOTA, 78.0\% MOTA, and 69.4\% IDF1 in visibility tracking.
The intracorporeal tracking is the most challenging since the major factors marking the entry and exit of the tools from the body are not readily visible. SurgiTrack however leverages the proposed direction-aware features, which could be linked to the surgeon operators hands and its rich fine-grained history to estimate the out-of-view and out-of-body status of the tools and outperforms all the existing methods with wide margin.
The intraoperative trajectory comes in the middle in terms of difficulty. While it may be challenging to ascertain the persistence of a trajectory after re-insertion, the class features are also helpful especially for tools of different categories. The direction features, however, has a better tendency of estimating the persistent identity of different tools of the same class with a +29.8\% and +29.9\% HOTA higher than similarity and appearance features respectively. 
Remarkably, SurgiTrack jointly handle the 3 trajectory perspectives unlike the compared methods, where we train separate model for each perspective.

\begin{figure*}[t]
    \centering    
        \includegraphics[width=0.95\linewidth]{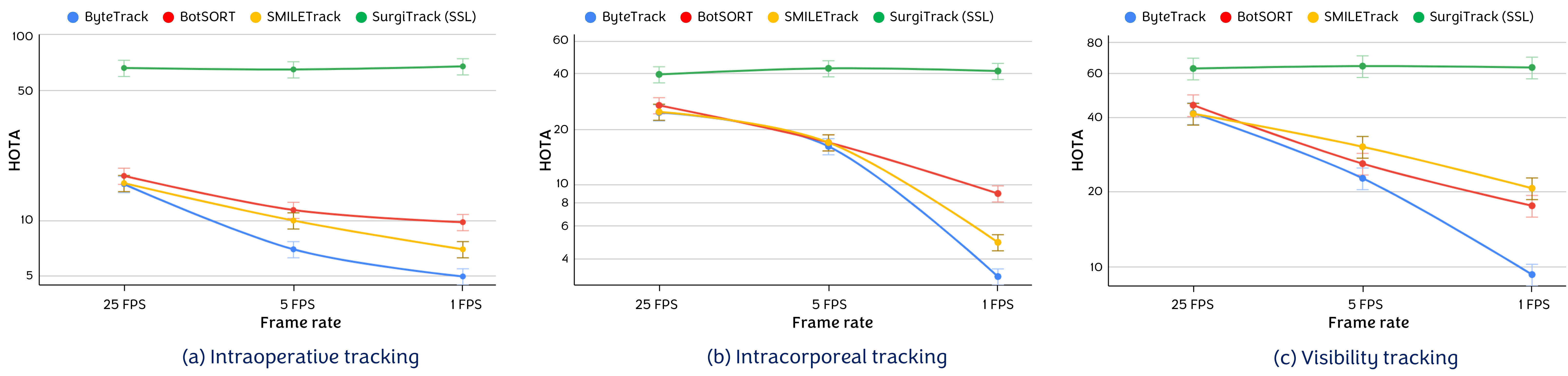}
        \caption{Impact of Direction Estimation in Tracking Surgical Tools at Varying Video Sampling Rates (i.e. 1, 5, 25 frames per seconds FPS). A demonstration is included in the qualitative video.}
        \label{fig:vfps}
\end{figure*}

\subsection{Analysis of Multi-Class Tracking Results}
We analyze in Table \ref{tab:classwise} the tracking performance per tool class and observe that hook is the best tracked with a HOTA above 75.6\% for the proposed solutions. 
This class-agnostic tracking result reveals that regularly used tools such as bipolar and clipper has a medium tracking accuracy. Rarely used tools (e.g. scissors, irrigator) have the least scores.
The grasper with the most multiple instances is very difficult to track, similarly the specimen bag is badly tracked due to its constantly deforming shapes. 

Revealed by this analysis, is the impact of the infusing surgical knowledge in the existing trackers, via the KB baselines. This greatly improves the tracking of multiple instances tools like grasper and marginal improves on tracking of single instance tools. Meanwhile, direction features learning the infusion of tool category and direction eliminates the need for hard-coded domain knowledge, giving a better performance.
With superior performance over appearance-based features in differentiating tools with identical appearances but different directions (e.g., graspers), we demonstrate that the learned features effectively capture direction-specific semantics, validating their designation as "direction" features.

\subsection{Impact of Direction Estimation on Tracking FPS}
We measure the impact of direction estimation on variable video sampling rates and show that it facilitates track re-identification in the events of non-overlapping bounding boxes as shown in Figure \ref{fig:vfps}. Unlike the compared models which are unable to track the tools at lower video sampling rates, SurgiTrack maintains a relatively stable tracking performance sampled across 1, 5, and 25 FPS in all the trajectory perspectives. This result justifies the re-identification strength of the proposed direction features and its modeling as a proxy to the tool operators, initially identified as the most promising cue for differentiating tool instances of the same class.

\begin{figure*}[!t]
    \centering
    \includegraphics[width=0.9\linewidth]{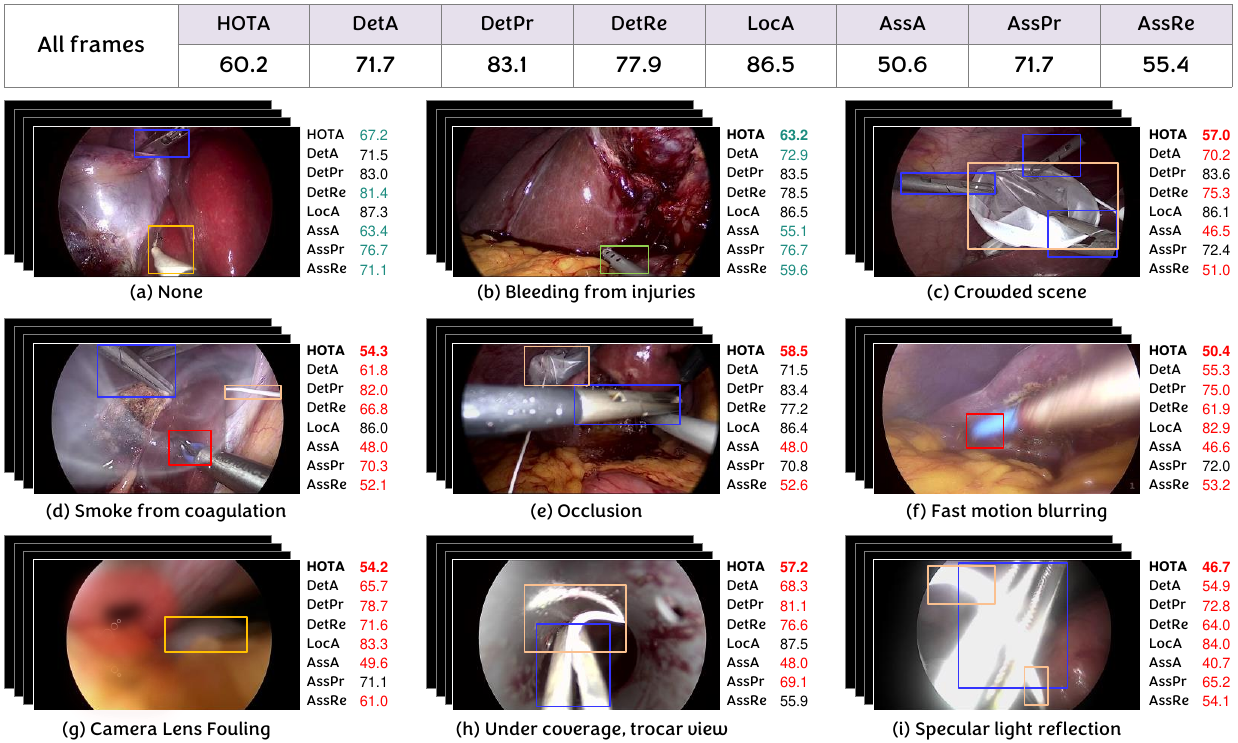}
    \caption{Performance Assessment of SurgiTrack Amidst Surgical Visual Challenges.
    Overall performance is tabulated at the top, preceded by quantitative and qualitative results showcasing tracking performance on specific visual challenge frames. Values in black denote comparable performance (within the average range, $\pm 1.0$). Values in green indicate above-average performance, while red values indicate decreasing performance below average. The breakdown explores distinct tracking metrics focusing on detection, localization, and association or re-identification. 
    A demo is included in the qualitative results video.}
    \label{fig:visual-challenge}
\end{figure*}

\subsection{Tracking Evaluation under Surgical Visual Challenges}
\label{sec:vc}

We analyze the tool tracking model across various surgical conditions, measured through HOTA metrics in Figure \ref{fig:visual-challenge}, to uncover insights into how the model interacts with the complexities of the surgical environment. 
The overall HOTA score across all frames for SurgiTrack (SSL) peaks at 60\%, providing a benchmark for comparison. Each visual condition is scrutinized based on its impact on different aspects of tracking, encompassing detection (accuracy, precision, recall), localization accuracy, and re-identification/association (accuracy, precision, recall).

Under clear scenes, devoid of visual challenges, the model excels with a substantial increase in performance (+7.0\% HOTA, +12.8\% AssA), showcasing efficient tool discernment, precise localization, and seamless tracking.
In scenarios with bleeding, the model demonstrates resilience, maintaining stable detection and localization, and showing improved association. This suggests the model's capability to detect and locate tools even amidst dynamic visual changes and potential tool discoloration from blood stains, which may have been treated as a form of augmentation owing to the abundant bleeding case samples.

Conversely, in crowded scenes, the model faces challenges as numerous tools closely packed together hinder detection accuracy. Disentangling cluttered tools proves difficult, potentially affecting accurate identity matching (-4.1\% AssA, -4.4\% AssRe).
Smoke from coagulation tools decreases the tracking score due to impaired visibility, resulting in significant missed detections (-9.9\% DetA, -11.1\% DetRe) and diminishing association accuracy.

\begin{figure*}[!tb]
    \centering
    \includegraphics[width=0.85\textwidth]{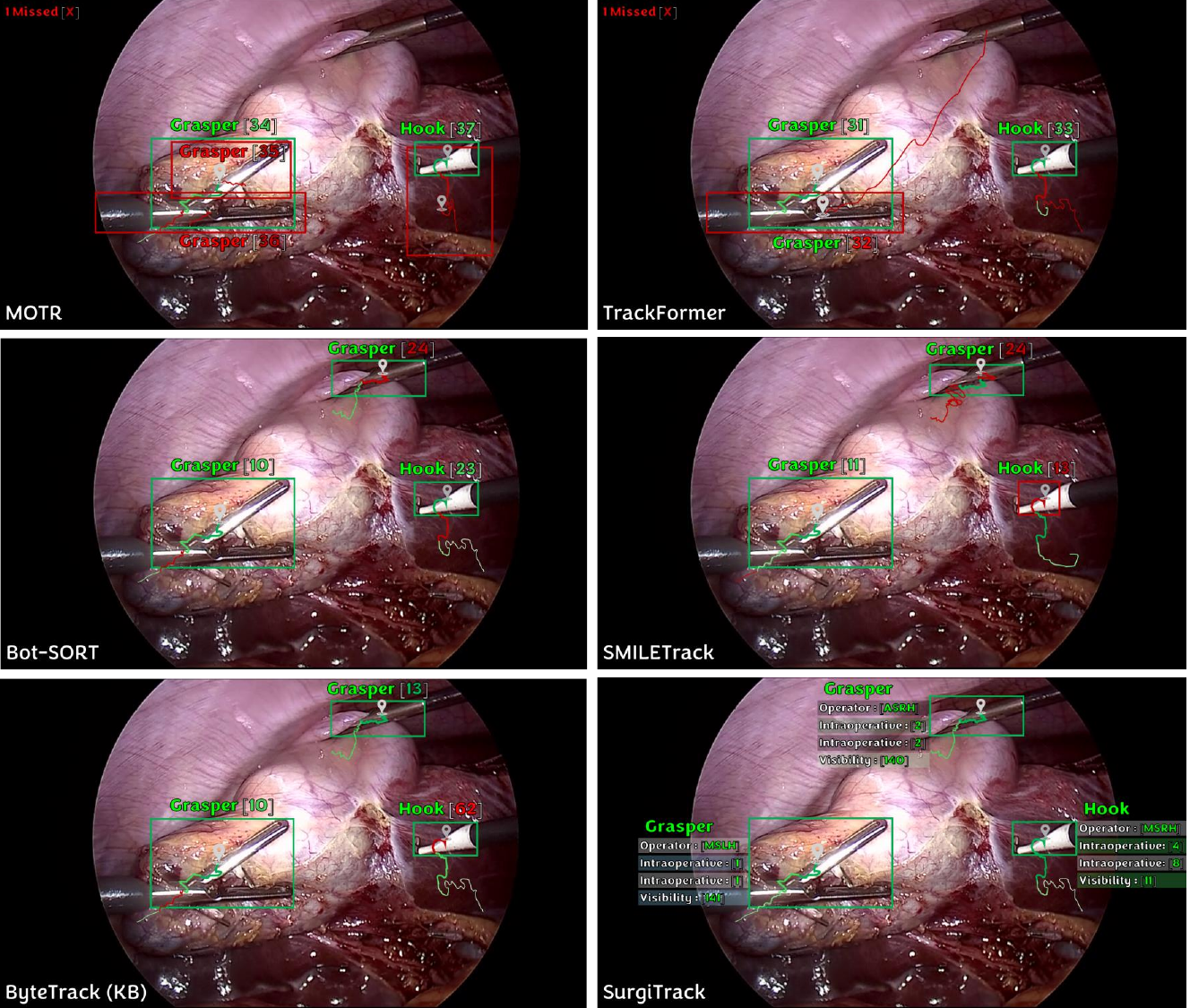}
    \caption{Qualitative result of SurgiTrack in comparison with some existing methods. Bounding box represents tool detection, tool name represents tool classification, number in block parenthesis represents track identity, and scribble represents tracklet (max. 2 seconds). Green color indicates correctness, red indicates failure. A demo is included in the qualitative results video.
    }
    \label{fig:qualitative-results-1}
\end{figure*}

\begin{figure*}[!tb]
    \centering
    \includegraphics[width=0.77\textwidth]{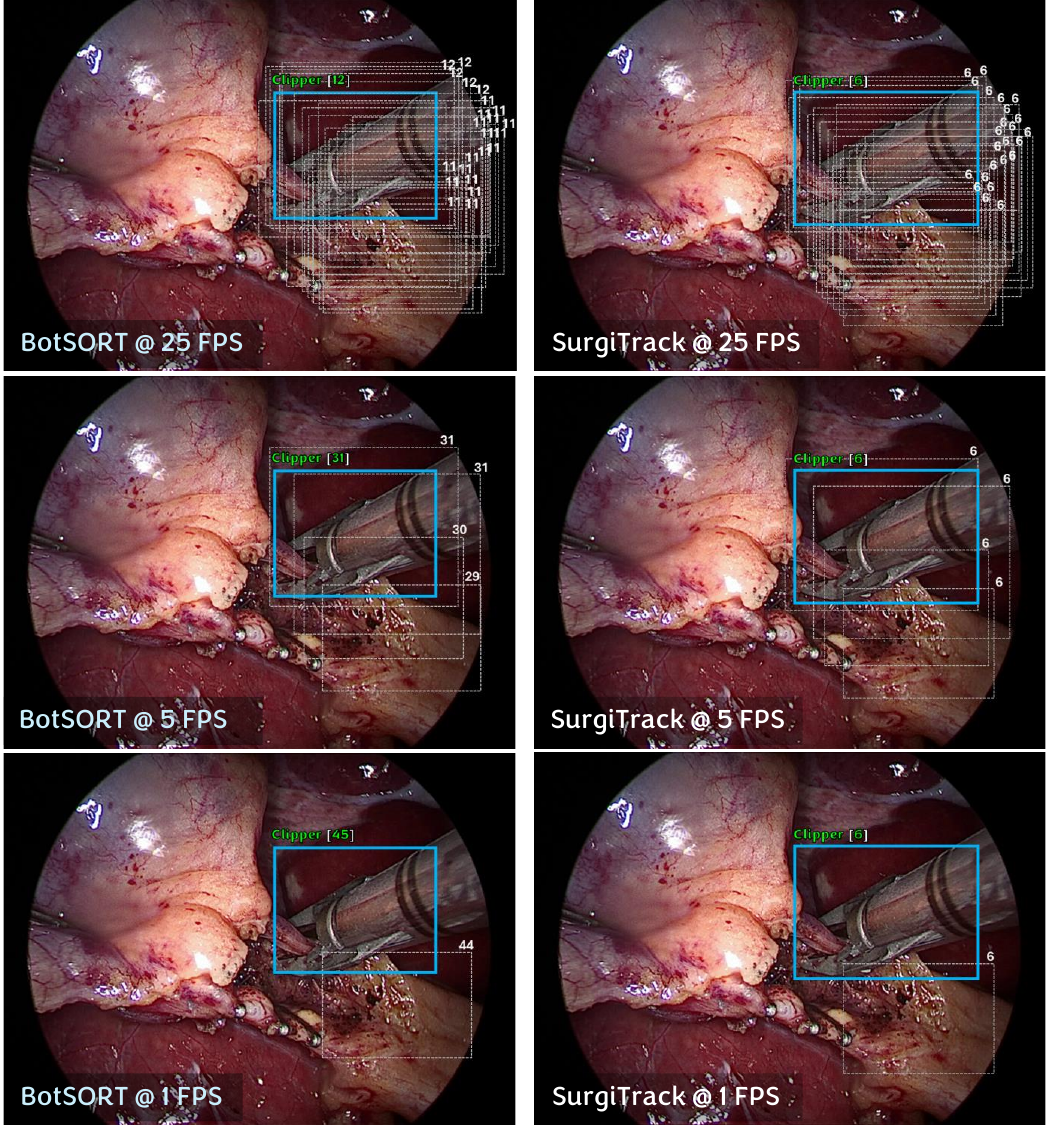}
    \caption{Qualitative result of SurgiTrack in comparison with a state of the art method (BotSORT) on tracking across variable frame rates (1FPS, 5FPS and 25FPS). Tick blue bounding box represents tool detection at current time. dotted gray bounding boxes detection at previous times, tool name represents tool classification, track identity number is written above each box. A demo is included in the qualitative results video.}
    \label{fig:qualitative-results-2}
\end{figure*}

The detection and localization accuracies remains unaffected in the cases of occlusion likely due to groundtruth bounding boxes including only the unoccluded tool regions. However, the occluded areas, which are often the discriminative tooltips, pose challenges for tool classification and re-identification tasks (-2.6\% AssA, -2.8\% AssRe), slightly impacting tracking performance.
Rapid motion-induced blurring introduces uncertainties in tool appearances, restricting the model's ability to accurately identify and track instruments, significantly reducing the HOTA score by -9.8\% and impacting all tracking aspects significantly.
Fluids such as blood and bile sometimes fouls the camera lens, impair visibility, and disrupt consistent identity matching. This negatively impact tracking in all aspects with a 6.0\% reduction in HOTA score. 

Limited camera coverage poses a major challenge, causing numerous missed detections, although localizing focused tools remains precise. Constrained trocar view hampers the model's ability to track and manage tool identities, especially complicating the interpretation of tool direction, as a link to the operating surgeon's hand.
Specular light reflection emerges as the most impactful visual challenge, leading to a notable decrease in all tracking aspects, with a huge loss of 13.5\% points in HOTA. The excessive brightness from specular light obscures visual details, leading to a 16.8\% and 9.9\% reduction in detection and association accuracies respectively.
{Though rare in this dataset, appearance changes in a single tool can also result from the use of electrocautery instruments.}

This analysis sheds light on how visual challenges during surgery impact vision tasks such as tool tracking. While some pose significant hurdles, others inadvertently offer clear visual cues that help track tools. The insights can inform future enhancements to improve tool tracking in challenging surgical environment.

Beyond the failure modes due to the surgical visual challenges, we observe that the model experiences difficulties at the beginning of a video due to limited historical data for the HGBM to rely on, leading to more frequent identity fragmentations, however, this usually stabilizes in few seconds.

As a limitation, our approach would not detect a replacement of a tool by another tool of the same class (and same manufacturer) if operated via the same trocar port. This is because the proposed direction feature is designed to uniquely differentiate tools based on their originating directions - the trocar ports. Additional signals would be needed to tackle this situation. The use of direction features may face challenges in certain type of surgery such as endoscopic endonasal surgery where the nostrils are closely space, or where trocars are closely spaced with little or no directional difference.

\subsection{Qualitative Results and Demos} 
In Figure \ref{fig:qualitative-results-1}, we present a qualitative analysis of our method alongside some of the existing approaches, complementing our quantitative evaluation. 
Performance-wise, transformer-based solutions (e.g., MOTR, TrackFormer) struggle with precise tool localization. Location-based methods (e.g., ByteTrack) mostly encounter identity mismatch issues. Appearance-based trackers (e.g., Bot-SORT) demonstrate strengths but sometimes misclassify tools and miss detections, akin to Similarity-based trackers such as SMILETrack.
Our method, SurgiTrack, displays enhanced resilience in tracking tools jointly across multiple perspectives compared to existing methods. 
Figure \ref{fig:qualitative-results-2} demonstrates SurgiTrack's robustness to variable frame rates (1, 5, 25 fps). This highlights the impact of direction re-ID features compared to the appearance features in the state-of-the-art method (BotSORT), which frequently assigns new track IDs when there is insufficient overlap between bounding boxes.

\emph{Video Demonstration}: Given that tracking is better appreciated in motion, as an appendix, we provide a full video demonstrating in detail the qualitative performance of our method across multiple evaluation settings.
The demo video is available at ~\videolink.

\section{Conclusion}
\label{sec:conclusion}
In this work, we propose, SurgiTrack, a novel deep learning approach for multi-class multi-tool tracking in surgical videos. 
Our approach utilizes an attention-based deep learning model for tool identity association by learning the tool motion direction which we conceived as a proxy to linking the tools to the operating surgeons' hands via the trocars.
We demonstrate that the motion direction features are superior to location, appearance, and similarity features for the re-identification of surgical tools given the non-distinctiveness of most tools' appearance, especially the ones from the same or similar classes. 
We show that the direction features can be learnt in 3 different paradigms of full-, weak-, and self-supervision depending on the availability of training labels.
We also design a harmonizing bipartite matching graph to enable non-conflicting and synchronized tracking of tools across three perspectives of intraoperative, intracorporeal, and visibility within the camera field of view, which represent the various ways of considering the temporal duration of a tool trajectory.
Additionally, we benchmark several deep learning methods for tool detection and tracking on the newly introduced CholecTrack20 dataset and conducted ablation studies on the suitability of existing re-identification features for accurate tool tracking.
Our proposed model emerges as a promising solution for multi-class multi-tool tracking in surgical procedures, showcasing adaptability across different training paradigms and demonstrating strong performance in essential tracking metrics. 
We also evaluate our model across different surgical visual challenges such as bleeding, smoke, occlusion, camera fouling, light reflection, etc., and presents insightful findings on their impact on visual tracking in surgical videos.
Qualitative results also show that our method is effective in handling challenging situations compare to the baselines and can effortlessly track tools irrespective of the video frame sampling rate.


\section*{Acknowledgements} \label{sec:thanks}
\noindent 
\small{
This work was supported by French state funds managed within the Plan Investissements d’Avenir by the ANR under references: National AI Chair AI4ORSafety [ANR-20-CHIA-0029-01], IHU Strasbourg [ANR-10-IAHU-02] and by BPI France [Project 5G-OR]. 
This work was granted access to the servers/HPC resources managed by CAMMA, IHU Strasbourg, Unistra Mesocentre, and GENCI-IDRIS [Grant 2021-AD011011638R3, 2021-AD011011638R4].
The authors acknowledge the suggestions of Engr. Rupak Bose on the visualization plots and thank Dr. Joël L. Lavanchy for some useful clinical discussions.}


\bibliographystyle{elsarticle-num}


\end{document}